\documentclass[journal, 10]{IEEEtran}
\IEEEoverridecommandlockouts
\usepackage{amsmath,amsfonts,amssymb}

\usepackage{times}
\usepackage{soul}
\usepackage{url}

\usepackage{cleveref}
\usepackage{setspace}
\usepackage{graphicx}
\usepackage{textcomp}
\usepackage{xcolor}
\usepackage[detect-all]{siunitx}

\usepackage{xspace}
\usepackage{subcaption}
\usepackage{bm}
\usepackage{array}

\usepackage{algorithm,tabularx}
\usepackage{algpseudocode}
\makeatletter
\newcommand{\multilines}[1]{%
	\begin{tabularx}{\dimexpr\linewidth-\ALG@thistlm}[t]{@{}X@{}}
		#1
	\end{tabularx}
}
\makeatother

\usepackage{mathtools}

\usepackage{amsthm}
\usepackage{multirow}
\usepackage{color}
\usepackage{epstopdf}
\usepackage{endnotes}
\usepackage{framed}
\usepackage{enumerate} 
\usepackage[subnum]{cases}
\usepackage{multicol}
\usepackage{tablefootnote}

\DeclarePairedDelimiterX{\norm}[1]{\lVert}{\rVert}{#1}
\DeclarePairedDelimiterX{\abs}[1]{\lvert}{\rvert}{#1}
\DeclarePairedDelimiterX{\innProd}[1]{\langle}{\rangle}{#1}
\DeclarePairedDelimiter\set{\lbrace}{\rbrace}

\newcommand{\SumNoLim}[2]{\ensuremath{\sum\nolimits_{#1}^{#2}}}
\newcommand{\SumLim}[2]{\ensuremath{\sum\limits_{#1}^{#2}}}

\newcommand{\nbigP}[1]{\ensuremath{(#1)}}

\newcommand{\bigP}[1]{\ensuremath{\bigl(#1\bigr)}}

\newcommand{\bigC}[1]{\ensuremath{\bigl\{#1\bigr\}}}
\newcommand{\biggP}[1]{\ensuremath{\biggl(#1\biggr)}}

\newcommand{\biggC}[1]{\ensuremath{\biggl\{#1\biggr\}}}
\newcommand{\BigP}[1]{\ensuremath{\Bigl(#1\Bigr)}}

\DeclareUnicodeCharacter{FB01}{fi}
\newcommand{\tbf}[1]{\textbf{#1}}

\newcommand{\centered}[1]{\begin{tabular}{l} #1 \end{tabular}}

\newcommand{\DD}[1]{\mathbb{#1}} 

\theoremstyle{plain}
\newtheorem{theorem}{Theorem}
\newtheorem{assumption}{Assumption}

\newtheorem{lemma}{Lemma}

\theoremstyle{definition}
\newtheorem{definition}{Definition}

\newcommand{\OurAlg}{{\small\textsf{DONE}}\xspace}

\AtBeginDocument{%
	\providecommand\BibTeX{{%
			\normalfont B\kern-0.5em{\scshape i\kern-0.25em b}\kern-0.8em\TeX}}}
\def\BibTeX{{\rm B\kern-.05em{\sc i\kern-.025em b}\kern-.08em
		T\kern-.1667em\lower.7ex\hbox{E}\kern-.125emX}}
\begin{document}

	\title{DONE: Distributed Approximate Newton-type Method for Federated Edge Learning}

	\author{
		Canh~T.~Dinh,
		Nguyen~H.~Tran\emph{, Senior Member, IEEE}, 
		Tuan Dung Nguyen, 
		Wei Bao\emph{, Member, IEEE},
		Amir Rezaei Balef,
		Bing B. Zhou\emph{, Member, IEEE},
		Albert Y. Zomaya\emph{, Fellow, IEEE}
		
		\IEEEcompsocitemizethanks{
				\IEEEcompsocthanksitem C.~T.~Dinh, N.~H.~Tran, W.~Bao, B.~B.~Zhou and A.Y.~Zomaya are with the School of Computer Science, The University of Sydney, Sydney, NSW 2006, Australia (email: \{canh.dinh, nguyen.tran, wei.bao, bing.zhou, albert.zomaya\}@sydney.edu.au). T. D. Nguyen is with the School of Computing, The Australian National University, Canberra, ACT 2601, Australia (e-mail: josh.nguyen@anu.edu.au). A. R. Balef is with Sharif University of Technology, Tehran, Iran (e-mail:amirbalef@gmail.com). \\
				This work is supported by the Vietnam National Foundation for Science and Technology Development (NAFOSTED) under Grant 102.02-2019.321. (Corresponding authors: Canh T. Dinh and Nguyen H. Tran) 
			
		}
	}
	
	
	\maketitle
	
	\begin{abstract}
		There is growing interest in applying distributed machine learning to edge computing, forming \emph{federated edge learning}. Federated edge learning faces non-i.i.d. and heterogeneous data, and the communication between edge workers, possibly through distant locations and with unstable wireless networks, is more costly than their local computational overhead. In this work, we propose \OurAlg, a distributed approximate Newton-type algorithm with fast convergence rate for communication-efficient federated edge learning. First, with strongly convex and smooth loss functions, \OurAlg approximates the Newton direction in a distributed manner using the classical Richardson iteration on each edge worker. Second, we prove that \OurAlg has linear-quadratic convergence and analyze its communication complexities. Finally, the experimental results with non-i.i.d. and heterogeneous data show that \OurAlg attains a comparable performance to Newton's method. Notably, \OurAlg requires fewer communication iterations compared to distributed gradient descent and outperforms DANE, FEDL, and GIANT, state-of-the-art approaches, in the case of non-quadratic loss functions.

	\end{abstract}
	
	\begin{IEEEkeywords}
		Distributed Machine Learning,  Federated Learning, Optimization Decomposition. 
	\end{IEEEkeywords}
	
	\IEEEpeerreviewmaketitle

	Traditional centralized machine learning procedures are gradually becoming inadequate as the computational and storage capacities of individual machines fall short of the data quantity involved in learning. The shift to intelligence at the edge \cite{parkWirelessNetworkIntelligence2019a,zhouEdgeIntelligencePaving2019a,wangConvergenceEdgeComputing2020}, including parallel and distributed approaches, are more capable, scalable and can be set up in various geographical locations. 

The existence of (i) powerful edge computing with data abundance and (ii) successful data center-type distributed machine learning architectures raises a natural question: \emph{Can we apply  large-scale distributed machine learning to edge computing networks}? We identify two key obstacles to overcome. {First, unlike the data sources of data center-type distributed machine learning, which are centrally collected, shuffled and hence homogenenous and i.i.d., 
data at edge networks are collected separately, and therefore heterogeneous and non-i.i.d., similar to cross-device federated learning \cite{konecnyFederatedLearningStrategies2017a,mcmahanCommunicationEfficientLearningDeep2017i,konecnyFederatedOptimizationDistributed2016a}; thus  distributed learning at edge networks is called \emph{federated edge learning}}. Second, we see that data center-type distributed machine learning  and  cross-device {federated learning} are two extremes: while the former involves learning with optimized computing nodes, data shuffling, and communication networks, the latter requires a massive number of participating devices with constrained computing, storage, and communication capacities. On the other hand, the storage and computing capacities of federated edge learning are comparable to the former, whereas its communication environments are similar to those of the latter due to notable physical distances between edge workers, multi-hop transmission, and different types of communication medium (e.g., wireless/optical/wireline/mmWave.) {Compared with local computation cost at each edge worker, the cost of edge communication is considerably higher in terms of  speed, delay,  and energy consumption \cite{tranFederatedLearningWireless2019d,wangConvergenceEdgeComputing2020,parkWirelessNetworkIntelligence2019a,tuNetworkAwareOptimizationDistributed2020, wangniGradientSparsificationCommunicationEfficient2017}; hence, it is often considered as a bottleneck specific to federated edge learning. }



Our goal is to develop a distributed algorithm which uses a minimal number of communication rounds to reach a certain precision for convergence and handles non-i.i.d. and heterogeneous data across edge workers.\footnote{{An important assumption in many machine learning methods is about independent and identically distributed (i.i.d.) data sample: all data observations are assumed to follow from the same probability distribution and are mutually independent. However, due to the nature of distributed learning in FL, this assumption may be violated (non-i.i.d. data).}} In the literature, first-order methods, which only use gradient information, can be robustly implemented in a distributed manner and require little local computation, are found most frequently \cite{bottouOptimizationMethodsLargeScale2018c}. Second-order counterparts like the Newton's method, which have also been well studied in literature \cite{boydConvexOptimization2004f,nesterovIntroductoryLecturesConvex2013b}, use  both gradient and curvature information. Their advantage is that by finding a ``better" descent direction, they can lead to  a substantially faster convergence rate, therefore requiring much fewer communication rounds to convergence.

In federated edge learning, the cost of local computation required by the (approximate) Newton-type method is gradually remedied by edge workers' improvements in computing power.  Although the Newton-type method is not friendly to distributed implementations due to its descent direction structure involving complicated inverse matrix-vector product, we design an algorithm that can overcome this issue and is summarized as follows.
\begin{itemize}
	\item We propose \OurAlg, a \underline{D}istributed appr\underline{O}ximate \underline{N}ewton-type m\underline{E}thod for federated edge learning. When the loss function of the learning task is strongly convex and smooth, \OurAlg  \emph{exploits the classical Richardson iteration} to enable the Newton's method to be distributively implementable on the edge workers. With large enough local rounds, \OurAlg generates an effective approximation to the Newton direction. In the literature, we are the first to approximate the Newton direction in a distributed setting using the Richardson iteration.
	\item Theoretically, we show that \OurAlg has a \emph{global linear-quadratic convergence rate}. We also analyze the computation and communication complexities of \OurAlg. Specifically,  with generalized linear models, the computation complexity of \OurAlg is comparable to that of the first-order methods. \OurAlg also never requires the explicit computation of the Hessian, but only needs the Hessian-vector product throughout, reducing the space requirement during optimization.
	\item Experimentally, in non-i.i.d. and heterogeneous settings,  we show that  \OurAlg achieves \emph{close performance of the true Newton's method in the same hyperparameter settings} and, therefore, significantly reduces the number of communication iterations compared to the standard first-order gradient descent (GD). \OurAlg also gains performance from DANE \cite{shamirCommunicationefficientDistributedOptimization2014a}, FEDL \cite{dinhFederatedLearningWireless2020a}, and GIANT \cite{wangGIANTGloballyImproved2018b},  the current popular distributed approximate Newton's algorithms for federated edge learning.
\end{itemize}
\section{Related Work} \label{Sec:RelatedWork}

\textbf{First-order distributed optimization methods.} First-order methods, which find a descent direction based on gradient information, are, paralellizable, easy to implement, and are the most common in practice. Stochastic gradient descent (SGD) \cite{zinkevichParallelizedStochasticGradient2010a}, 
accelerated SGD \cite{shamirDistributedStochasticOptimization2014a}, variance reduction SGD \cite{leeDistributedStochasticVariance2017a,reddiAIDEFastCommunication2016g}, stochastic coordinate descent
\cite{richtarikDistributedCoordinateDescent2016a}, 
and dual coordinate ascent \cite{yangTradingComputationCommunication2013a} 
are examples of this type of algorithms. 
The trade-off for less computation at the edge, i.e., finding only the gradient and possibly some additional information, is the requirement of many iterations until convergence. Communication overheads become a bigger problem in multi-hop edge networks where the bandwidth varies or is limited among workers. {\cite{shiDeviceSchedulingFast2019, yangSchedulingPoliciesFederated2020} apply scheduling policies while \cite{tangDoubleSqueezeParallelStochastic2019} uses compression techniques to alleviate the communication pressure, but they base on the first-order gradient method.}

\textbf{Newton-type distributed optimization methods.} Different from first-order methods,  Newton-type methods use not only gradient but also curvature information to find a descent direction. Existing distributed Newton-type algorithms  include DANE \cite{shamirCommunicationefficientDistributedOptimization2014a}, AIDE \cite{reddiAIDEFastCommunication2016g}, DiSCO \cite{zhangDiSCODistributedOptimization2015a}, and GIANT \cite{wangGIANTGloballyImproved2018b}.   DANE requires each worker to solve a well-designed local optimization problem, whereas  GIANT \cite{wangGIANTGloballyImproved2018b} uses the harmonic mean (instead of the true arithmetic mean) of the Hessian matrices. On the other hand, DiSCO is an inexact damped Newton method using a preconditioned conjugate gradient algorithm, which requires multiple communication exchanges between the workers to find the Newton's direction. Our proposed \OurAlg  is another type of  distributed \emph{approximate} Newton method that is based on Richardson's iteration for local update. 

\textbf{Federated edge learning.} Thanks to the requirement of real-time processing of data in many applications and the improvement in computing power and storage capacity of smart devices, research into intelligence at the edge has proliferated in recent years \cite{zhouEdgeIntelligencePaving2019a}. An example of cross-device federated learning algorithm that has been studied extensively is {FedAvg}  \cite{mcmahanCommunicationEfficientLearningDeep2017i}, which includes a large network of highly diverse devices participating in learning a global predictive model. On the other hand, one of the early adopters of federated edge learning is \cite{wangAdaptiveFederatedLearning2019b}, who propose {the use of GD} for a FedAvg-type algorithm. In such distributed computing methods, the cost of communication is generally significant and poses serious problems in operation, especially in networks where limited bandwidth and high latency are prominent. 	{\cite{konecnyFederatedLearningStrategies2017a} proposes a method to reduce the uplink communication costs; however, this method requires uploading the Hessian matrix to the server for aggregation.} {\cite{dinhFederatedLearningWireless2020a} proposes FEDL, a first-order method for federated learning and shows that there is always a trade-off between computation and communication: to require less communication during training, the amount of computational processing in edge workers/devices has to increase.} 
\section{\OurAlg: Distributed Approximate Newton-type Method for Federated Edge Learning} \label{Sec:Algorithm}
\subsection{Optimization Problem}
In the context of federated edge learning, there are $n$ edge workers, located at different sites and  communicating with an edge aggregator\footnote{The roles of edge workers and the aggregator are respectively similar to  worker nodes and parameter server in data center-type distributed machine learning. } to learn a model $w^*$ which is a solution to the following problem
\begin{align}
	\min_{w \in \mathbb{R}^d} f (w) \doteq \frac{1}{n} \SumLim{i =1}{n} f_i ( w),  \label{E:Global_Loss}
\end{align}
where $f_i ( w)$ is the loss function of worker $i$. Each worker $i$  has a local dataset containing a collection of $D_i$  samples  $\{a_j, y_j\}_{j=1}^{D_i}$, where $a_j \in \mathbb{R}^d$ is an input and $y_j \in \mathbb{R}$ can be a target response (or label). Then the (regularized) loss function of each worker $i$ is the average of  losses on its data points\footnote{In machine learning, this problem is called the empirical risk minimization (ERM).} as follows
\begin{align}
	{f_i}(w) = \frac{1}{{D_i}} \SumLim{j=1}{D_i}  { l(w, (a_j, y_j)) + \frac{\lambda}{2} \norm{w}^2}, \label{E:GLM}
\end{align}
where the regularization term $\frac{\lambda}{2} \norm{w}^2$ is added to improve stability and generalization. Throughout this paper, we use $\norm{\cdot}$ as the Euclidean norm for vectors, and spectral norm for matrices.  Some examples of the loss function are:  linear regression  with $ l(w, (a_j, y_j))  = \frac{1}{2} (\innProd{a_j, w} - y_j)^2,  y_j \in \DD{R}$, and  logistic regression with  $l(w, (a_j, y_j))   = \log \bigP{ 1 + \exp ( - y_j \innProd{a_j, w}) }, y_j \in \{-1, 1\}$, where $\langle x, y\rangle$ denotes the inner product of vectors $x$ and $y$. 

\begin{assumption} \label{Assumption}
	The function $f:\mathbb{R}^d \rightarrow \mathbb{R}$ is twice continuously differentiable, $L$-smooth, and $\lambda$-strongly convex, $\lambda > 0$, i.e., 
	\begin{align*}
		\lambda I \preceq \nabla^2 f(w) \preceq  L I, \quad \forall w \in \mathbb{R}^d, 
	\end{align*}
	where $\nabla^2 f(w) \in \mathbb{R}^{d \times d}$ and $I \in \mathbb{R}^{d \times d}$ denote the Hessian of $f$ at $w$ and identity matrix, respectively. 
\end{assumption}
We use Assumption~\ref{Assumption} throughout this paper. We note that strong convexity and smoothness in Assumption~\ref{Assumption} can be found in a wide range of loss functions such as linear regression and logistic regression as above. We also denote $ \kappa  \doteq \frac{L}{\lambda}$ the condition number of  $f$, where large $\kappa$ means $f$ is ill-conditioned, which may require high computational complexity to optimize. 

\vspace{-0.4cm}
\subsection{Challenges to Distributed Newton's Method}
We first review the Newton's method \cite{boydConvexOptimization2004f}  to solve \eqref{E:Global_Loss} as follows
\begin{align}
	w_{t+1}&= w_{t} - \BigP{\nabla^2 f(w_t)}^{-1} \nabla {f}(w_t) \nonumber\\
	&= w_{t} - \BigP{\frac{1}{n}\SumLim{i=1}{n} \nabla^2 f_i(w_t)}^{-1} \BigP{\frac{1}{n}\SumLim{i=1}{n} \nabla f_i(w_t)}. \label{E:Newton}
\end{align}

{We denote the \emph{global} Newton direction in iteration $t$ as $d_t \doteq -\bigP{\nabla^2 f(w_t)}^{-1} \nabla {f}(w_t)$. In the vanilla Newton's method, the update rule \eqref{E:Newton} becomes $w_{t+1} = w_t + d_t$.}

One might be tempted to allow each worker to send $\nabla {f_i}(w_t)$ and $\nabla^2 f_i(w_t)$ to the aggregator for updating \eqref{E:Newton}. However, the crux to designing a distributed Newton's method is \emph{Hessian-free communication} and \emph{inverse-Hessian-free computation}. Indeed, sending Hessians of size $O(d^2)$ over the network or computing the inverse Hessian with complexity $O(\SumNoLim{i=1}{n}D_i d^2 + d^3)$, where $d$ is the dimension of data,  is considered impractical considering high-dimensional feature vector or ``big data'' size. 

Some approaches such as in \cite{shamirCommunicationefficientDistributedOptimization2014a,wangGIANTGloballyImproved2018b} allow each worker to approximately calculate $\bigP{\nabla^2 f_i (w_t)}^{-1} \nabla {f}(w_t)$ and send this vector to the edge aggregator for the following update
\begin{align*}
	w_{t+1} = w_{t} - \frac{1}{n}\SumLim{i=1}{n} \bigP{ \nabla^2 f_i(w_t)}^{-1} \nabla {f}(w_t). 
\end{align*}

We next review the key ingredient that helps \OurAlg overcome these challenges. 
\subsection{Richardson Iteration Review}

{The Newton direction $d_t$ is the solution to the following system of linear equations:
\begin{align*}
\nabla^2 f(w_t) d_t = -\nabla {f}(w_t).
\end{align*}
As mentioned earlier, the cost of exactly solving this equation is $O(d^3)$, which can be impractical if the model is high-dimensional. Here we describe the Richardson iteration, a method to solve this approximately but at a lower cost.
}

The purpose of the Richardson iteration is to find the vector $x^*$ satisfying the linear system $A x^* = b$, assuming $A\in \mathbb{R}^{d \times d}$ is a symmetric and positive definite matrix. 
\begin{align*}
	x_k =  \nbigP{I - \alpha A}  x_{k-1} + \alpha b, \quad k =  1, 2, \ldots
\end{align*}
Let $\lambda_{max}(A)$ and $\lambda_{min}(A)$ denote the largest and smallest eigenvalues of $A$, respectively. The Richardson iteration converges, i.e., $\lim_{k \rightarrow \infty} x_k = x^* = A^{-1} b$, if and only if 
\begin{align}
	0 < \alpha < \frac{2}{\lambda_{max}(A)}, \label{E:alpha_convergence}
\end{align}
which ensures $\norm{I - \alpha A} < 1$. 
The details of Richardson iteration analysis can be found in \cite{rheinboldtClassicalIterativeMethodsa}. From another viewpoint, the Richardson iteration is equivalent to using the GD method to solve the following quadratic problem
\begin{align*}
	\min_{y \in \mathbb{R}^d} \quad \frac{1}{2}\innProd{y, Ay} - \innProd{y, b}. 
\end{align*}
\subsection{\OurAlg Algorithm }

The general idea of \OurAlg, as presented in in Alg.~\ref{Alg:1}, is each client finds its local Newton direction vector, and the server will aggregate these local vectors to approximate the global Newton directions. We see that $2T$ is the number of communication iterations between the aggregator and edge workers. Note that the exact global gradient $\nabla {f}(w_t)$ is required (line~\ref{line:grad}), which is the key to the convergence of \OurAlg to be shown later. Define the \emph{local} Newton direction on client $i$ in iteration $t$ as $d_{i, t} \doteq \bigP{\nabla^2 f_i(w_t)}^{-1} \nabla {f_i}(w_t)$. To find the local Newton direction $d_{i, t}$, Alg.~\ref{Alg:1} uses $R$ Richardson iterations, resulting in the \emph{approximate} local Newton direction $d_{i, t}^{R}$ (line~\ref{line:local_model}). 

\textbf{The gist of \OurAlg}: Even though each worker uses the Richardson iteration to obtain its local Newton direction $d_{i, t}^{R}$, their average $d_{t}^{R}$ at the aggregator well approximates the global Newton direction when $\alpha \leq 1/R$. This is demonstrated in the following result.  

\begin{theorem} \label{Th:1}
	Consider symmetric and positive definite matrices $A_i \in \mathbb{R}^{d \times d}, i=1, \ldots, n$. Let $A = \frac{1}{n}\SumNoLim{i=1}{n} A_i$, and $x_k$ and $x_{i,k}$ follow the Richardson iteration as follows
	\begin{align*}
		x_k &=  \nbigP{I - \alpha A}  x_{k-1} + \alpha b \\
		x_{i,k} &=  \nbigP{I - \alpha A_i}  x_{i,k-1} + \alpha b, \quad  k=1, 2,  \dots
	\end{align*}
	with $0< \alpha < \frac{2}{\lambda_{max}(A)}$, then we have
	\begin{align*}
		\lim_{k \rightarrow \infty}x_k = A^{-1} b = x^*, \quad \lim_{k \rightarrow \infty}  x_{i,k} = A_i^{-1} b.
	\end{align*}
	Especially, when $x_{i,0} = x_{0}$ and $\alpha \leq \min \bigC{\frac{1}{k}, \frac{1}{\hat{\lambda}_{max}}}$, where $\hat{\lambda}_{max} \doteq \max_i \lambda_{max}(A_i)$, we have
	\begin{align}
		\norm*{\frac{1}{n}\SumLim{i=1}{n}x_{i,k} - x^*} &\leq E_1 + E_2, \nonumber
	\end{align}
	where 
	\begin{align*}
		E_1 &= \norm*{x_k - x^*} \leq \norm*{(I - \alpha A)^k} \norm*{x_0 - x^*}, \nonumber \\
		E_2 &= \norm*{\frac{1}{n}\SumLim{i=1}{n}x_{i,k} - x_k} \leq O\biggP{ \frac{\nu}{k^2}  \bigP{\norm*{b} + \norm*{x_0}}}, \nonumber \\
		 \nu &=  \norm*{A^2 -	\frac{1}{n}\SumLim{i=1}{n} A_i^2}.
	\end{align*}
	
\end{theorem}

{The proof to Thm.~\ref{Th:1} can be found in \ref{Proof:Th:1}}. Observing that $d_{i, t}^{r}, \nabla^2 f_i(w_t),$ and $-\nabla {f}(w_t)$ of \OurAlg play the same roles as $x_{i,k}, A_i,$ and $b$ in the above theorem, respectively, we obtain the following remarks.

\begin{itemize}
	\item The only parameter of \OurAlg to be fine-tuned is $\alpha$. As it is impractical to compute the eigenvalues of $\nabla^2f_i(w_t)$ to guarantee convergence as in \eqref{E:alpha_convergence}
	, the {Richardson iteration} may not converge. In practice, however, a sufficiently small $\alpha$ almost always works, but very small $\alpha$ can lead to slow convergence (c.f. Section~\ref{Sec:Sim}). 
	
	
	\item The theory of \OurAlg requires full local data passing and full participation of all edge workers. However, using mini-batches is common in machine learning to reduce the computation bottleneck at the worker. Another critical issue is the straggler effect, in which the run-time in each iteration  is limited by the slowest worker (the straggler) because  heterogeneous edge workers compute and communicate at different speeds. Thus, choosing a subset of participating workers in the aggregation phase is a practical approach to reduce the straggler effect. In Section~\ref{Sec:Sim}, we will experimentally show that  \OurAlg works well with mini-batches and worker subset sampling\footnote{By uniformly at random choosing a subset of $B$ data points from $D_i$ samples and a subset of  $S \leq n$ workers, respectively.} when approximating the Hessians $\nabla^2 f_i(w_t)$ and mitigating the straggler effect, respectively. 
	\item It is obvious that the larger $R$ and $T$, the higher computation and communication complexities of \OurAlg, yet the better accuracy and smaller optimal gap, for the  Newton direction and convergence, respectively. In the next section, we will quantify how much $R$ and $T$ affect the convergence of \OurAlg. 
	\item In Theorem \ref{Th:1}, the error of the approximation is quantified by two terms, $E_1$ and $E_2$. $E_1$ is the error between the \emph{centralized} Richardson iterate and the solution $x^*$; in the context of \OurAlg, this is the distance between the approximate Newton direction (if all Hessians are sent to the server) and the true Newton direction. On the other hand, $E_2$ is the error between the \emph{centralized} iterate and the average of \emph{decentralized} iterates. Therefore, this distributed approximation incurs an additional error term $E_2$. Also, $E_2$ depends primarily on  $\nu = \norm*{A^2 - \frac{1}{n} \SumNoLim{i=1}{n} A_i^2}$, which quantifies the heterogeneity in edges' data. In other words, if statistical heterogeneity is more significant, each edge must run more Richardson iterations to approximate the Newton direction to the same precision. Finally, we observe a relationship between $\alpha$ and $k$ (or $R$ in the case of \OurAlg) here: to distributively approximate the Newton direction, $\alpha$ must decrease as $k$ (or $R$) increases to avoid divergence of the average Richardson iterate.
\end{itemize}

\begin{algorithm} [t] 
	\caption{\OurAlg} \label{Alg:1}
	\begin{algorithmic}[1]
		\State \tbf{input:} $T$, $R$, $\alpha$, $w_0$
		\For {$t =  0$ to  $T-1$} 
		\State Aggregator  sends $w_t$ to all edge workers 
		\For{\text{all edge workers $i = 1, \ldots, n$ in parallel}}
		\State \multilines{Send $\nabla {f_i}(w_t)$ to the edge aggregator, then receives  the  aggregated gradient $\nabla {f}(w_t) = \frac{1}{n} \SumNoLim{i = 1}{n}\nabla {f_i}(w_t)$ sent back by the edge aggregator. } \label{line:grad}
		\State Set $d_{i,t}^0 = 0$ 
		\For {$r =  1 \text{\,to\,}  R$} 
		\State 	$d_{i, t}^{r} = \bigP{I - \alpha \nabla^2 f_i (w_t)} d_{i, t}^{r-1} -  \alpha \nabla {f}(w_t) $		\label{line:local_model}
		\EndFor	
		\State Send $d_{i, t}^{R}$ to the aggregator
		\EndFor
		\State Aggregator receives $d_{i, t}^{R}$ from all workers and  updates: 
		\begin{align}
			w_{t+1} &= w_{t} + \eta_t d_t^{R}, 
			\text{where  } d_t^R \doteq \frac{1}{n}\SumNoLim{i=1}{n} d_{i, t}^{R}. \label{line:glob} 
		\end{align}
		\EndFor
	\end{algorithmic}
\end{algorithm}

\section{Convergence and Complexity Analysis} \label{Sec:Convergence}
In this section, we will provide the convergence  and complexity analysis of \OurAlg, and compare it with distributed GD, DANE \cite{shamirCommunicationefficientDistributedOptimization2014a}, and FEDL \cite{dinhFederatedLearningWireless2020a}. 
\subsection{Convergence analysis}

We see that each edge worker has $R$ computation rounds, and the  complexity at each round is calculating $d_{i, t}^{r}$ on line~\ref{line:local_model}. It is straightforward to see that the bottleneck of this step is performing the matrix-vector product with $O(D_i \cdot d^2)$ computation complexity. However, this complexity can be reduced to $O(D_i \cdot d)$  with a special class of generalized linear models (GLM) \cite{hastieElementsStatisticalLearning2009a} having the linear term $\innProd{a_j, w}$ in its loss function,  e.g., regularized linear regression or logistic regression. Indeed, from \eqref{E:GLM}, the Hessian of $f_i$ in these learning tasks is 
\begin{align*}
	\nabla^2 {f_i}(w_{t}) = \frac{1}{{D_i}} \SumLim{j=1}{D_i}  \beta a_j a_j^T + \lambda I, 
\end{align*}
where $\beta$ is a scalar that depends on $\innProd{a_j, w_t}$. Therefore, the matrix-vector product $\nabla^2 f_i (w_t) d_{i, t}^{r-1}$ becomes 
\begin{align*}
	\frac{1}{{D_i}} \SumLim{j=1}{D_i}  \beta a_j \innProd{a_j, d_{i, t}^{r-1}}, 
\end{align*} 
which contains only vector-vector products with computation complexity $O(D_i \cdot d)$. Therefore,  with GLM, the {computation complexity of each worker $i$} using \OurAlg is $O(D_i\cdot d \cdot R \cdot T)$, which only scales linearly with respect to (w.r.t.) the data size and feature dimension. 

Before analyzing the effects of $R$ and $T$ on \OurAlg's convergence, we need the following standard assumption for the Newton's method analysis \cite{boydConvexOptimization2004f}. 
\begin{assumption} \label{Asm:2}
	The Hessian of $f$ is $M$-Lipschitz continuous, i.e.,   $\norm{\nabla^2 f(w)  - \nabla^2 f (w')}  \leq M \norm{w - w'}, \, \forall w, w' \in \mathbb{R}^d.$
\end{assumption}
	Here, the value of $M$ measures how well $f$ can be approximated by a quadratic function, e.g., $M=0$ for when f is quadratic such as in the linear regression case. In this work, we  focus on the global convergence of \OurAlg without using the backtracking line search. In order to enable the global convergence of the Newton-type method, we choose an adaptive stepsize of $\eta_t$ introduced in \cite{polyakNewVersionsNewton2020a}:
	\begin{align}
		\eta_t = \min \biggC{1,\frac{ \lambda ^2}{L \norm{\nabla {f(w_t)}}}}. \label{E:eta}
	\end{align}	
	In the damped Newton phase, the edge aggregator will update $w_{t+1} = w_{t} +  \eta_t d_t^{R}$ with an adaptive step size $\eta_t = \frac{ \lambda ^2}{L \norm{\nabla {f(w_t)}}}$ to ensure $d_t^R$ is a descent direction. When the damped Newton phase finishes, $\eta_t=1$ and the algorithm enters the pure Newton phase.
For practical \OurAlg, we need to choose a finite value for $R$, which means that $d_t^R$ is an approximation of the Newton direction.  We next define a parameter $\delta$ that measures how well $d_t^R$ approximates the true Newton direction. In the sequel, for the ease of presentation, we  denote $g_t \doteq \nabla f (w_t), H_t \doteq  \nabla^2 f (w_t),$ and $\hat{d}_t \doteq - H_t^{-1} g_t$. 
\begin{definition} \label{Def1}
	$d_t^R$ is called a $\delta$-approximate $\hat{d}_t$ if
	\begin{align*}
		\norm{ \hat{d}_t - d_{t}^{R} }  \leq \delta  \norm{\hat{d}_t}. 
	\end{align*}
\end{definition}
In other words, $\delta$ captures the inexact level of the approximate solution  $d_{t}^{R}$ to the equation $H_t x^*= -g_t$ with the true solution $x^* = \hat{d}_t$. 

The following result shows the relationship between $R$ and $\delta$, and how the $\delta$-approximation affects the convergence rate of \OurAlg. 
	\begin{lemma} \label{Lem:2}
		Let Assumptions~\ref{Assumption} and \ref{Asm:2} hold. For an arbitrary small  $\alpha \leq \min \bigC{\frac{1}{R}, \frac{1}{\hat{\lambda}_{max}}}$, if we set  
		\begin{align*}
			\delta = \norm*{(I - \alpha A)^R}   + O\BigP{\frac{\nu L}{R^2} },\\
			t_0 = \max \biggC{0, \bigg \lceil \frac{2L}{\lambda^2\norm{\nabla {f(w_0)}}} \bigg \rceil - 2},\\ \nonumber
			\gamma =  \frac{L}{2\lambda^2}\norm{\nabla {f(w_0)}} - \frac{t_0}{4} \in \big[ 0, \frac{1}{2}\big);  \nonumber
		\end{align*}
		then \OurAlg has 
		\begin{align}
			\norm{w_{t} - w^*} \leq \begin{cases}
				\frac{1}{\kappa} (t_0 - t + \frac{2\gamma}{1-\gamma}) +  \frac{\delta}{\kappa} , t \leq t_0 \label{E:lem2}\\ 
				\frac{2t\gamma^{2^{t-t_0}}}{\kappa(1 - \gamma^{2^{t-t_0}})} +  {(\delta\kappa)}^t \norm{w_{0} -w^* },  t > t_0. 
			\end{cases}
		\end{align}
	\end{lemma}
	
	{The proof can be found in \ref{Proof:Lem:2}}.  Lemma~\ref{Lem:2} shows that in damped phase, $ \norm{w_{t} - w^* }$ is decreased by at least a constant at each iteration. In pure Newton phase, \OurAlg has a linear-quadratic convergence rate. The quadratic term in this lemma is  equivalent to \cite{polyakNewVersionsNewton2020a}, whereas the linear term is due to the $\delta$-approximation of Newton direction of \OurAlg. The $\delta$-approximation includes two terms. The first term comes from using Richarsion Interaction to approximate true Newton direction and the second one depends on the heterogeneity in edges' data. It is obvious that when $R$ increases, $\delta$ decreases, and thus the linear term disappears, recovering the quadratic convergence of the adaptive Newton method. Lemma~\ref{Lem:2} also shows when $\kappa$ is large, i.e., more computation rounds are needed to reduce the approximation error $\delta$ to guarantee $\delta\kappa < 1$.
\subsection{Complexity analysis}
\begin{table}[t!]
	\small
	\setlength{\tabcolsep}{2pt}
	\centering
	
	\caption{ Communication and computation complexity comparison for the ridge regression problem. \OurAlg achieves a global linear or quadratic convergence depending on the regime specified in Theorems \ref{Th:2} and \ref{Th:3}. The computation complexity of \OurAlg with GLM is comparable to that of the first-order methods. GIANT is not considered for the comparison as it only has a local convergence rate.}
	\vspace{-1mm}
	\begin{tabular}{|c|c|c|}
		\hline
		\centered{\textbf{Method}} & \centered{\textbf{Communication}} & \multicolumn{1}{c|}{\textbf{Computation}} \\ \hline
		\multicolumn{1}{|c|}{\multirow{2}{*}{\OurAlg}} & \multicolumn{1}{c|}{$O\BigP{\delta \kappa \log \frac{1}{\epsilon}}$}    & \multicolumn{1}{c|}{${O}\BigP{D_i \cdot d \cdot R \cdot \delta \kappa \log \frac{1}{\epsilon}}$} \\ \cline{2-3}
		\multicolumn{1}{|c|}{} & \multicolumn{1}{c|}{$O\BigP{\log \log \frac{1}{\epsilon}}$}        & \multicolumn{1}{c|}{${O}\BigP{D_i \cdot d \cdot R \cdot \log \log \frac{1}{\epsilon}}$} \\\hline
		\centered{DANE}  & \multicolumn{1}{c|}{${O}\BigP{\frac{\kappa^2}{D_i} \log (d n)\log \frac{1}{\epsilon}}$}            & \multicolumn{1}{c|}{${O}\BigP{d \cdot R \cdot \kappa^2 \log (d n)\log \frac{1}{\epsilon}}$} \\ \hline
		\centered{FEDL} & \multicolumn{1}{c|}{$O\BigP{\frac{1}{\Theta}\log \frac{1}{\epsilon}}$} & \multicolumn{1}{c|}{${O}\BigP{D_i \cdot d \cdot R \cdot \frac{1}{\Theta} \log \frac{1}{\epsilon}}$} \\ \hline
		\centered{GD} & \multicolumn{1}{c|}{$O\BigP{\kappa\log \frac{1}{\epsilon}}$} & \multicolumn{1}{c|}{${O}\BigP{D_i \cdot d \cdot \kappa\log \frac{1}{\epsilon}}$} \\ \hline
	\end{tabular}
	\label{T:compare}	
	\vspace{-3mm}
\end{table}
We next address the communication complexity $O(T)$ for the global convergence of \OurAlg to $w^*$. 
\begin{definition} \label{Def2}
	We define $w_T$ an $\epsilon$-optimal solution to \eqref{E:Global_Loss} if 
	\begin{align*}
		\norm{w_{T} - w^*}  \leq \epsilon. 
	\end{align*}
\end{definition}
	From Lemma~\ref{Lem:2}, we see that there are regimes  in which \OurAlg has linear or quadratic convergence rate, and these regimes depend on the $\delta$ setting and the initialization $w_0$ in a neighborhood of $w^*$. In the following we will identify two extreme regimes. 
	
	In the first regime where the $\delta$-approximation error dominates the bound in \eqref{E:lem2}, i.e., 
	$\norm{w_{0} -w^* } \geq \frac{2t\gamma^{2^{t-t_0}}}{\kappa  {(\delta\kappa)}^t(1 - \gamma^{2^{t-t_0}})}  $
	then \OurAlg has linear convergence. 
	
	\begin{theorem} \label{Th:2}
		If $\norm{w_{0} -w^* } \geq \frac{2t\gamma^{2^{t-t_0}}}{\kappa  {(\delta\kappa)}^t(1 - \gamma^{2^{t-t_0}})}$, and assume that $\delta\kappa <1$, then \OurAlg has linear convergence
		\begin{align*}
			\norm{w_{t} - w^*}  \leq  2{(\delta\kappa)}^t \norm{w_{0} -w^* },  
		\end{align*}
		and  its communication complexity is
		\begin{align}
			T = O \BigP{\delta\kappa\log \frac{1}{\epsilon}}. \label{E:Th21}
		\end{align}
	\end{theorem}
	{The proof can be found in \ref{Proof:Th:2}.} In the second regime where the $\delta$-approximation error is negligible, i.e., 
	$\norm{w_{0} -w^* } < \frac{2t\gamma^{2^{t-t_0}}}{\kappa  {(\delta\kappa)}^t(1 - \gamma^{2^{t-t_0}})}  $
	, then \OurAlg has  quadratic convergence.

	\begin{theorem} \label{Th:3}
		If 
		$\norm{w_{0} -w^* } < \frac{2t\gamma^{2^{t-t_0}}}{\kappa  {(\delta\kappa)}^t(1 - \gamma^{2^{t-t_0}})}  $
		then \OurAlg has quadratic convergence
		\begin{align*}
			\norm{w_{t} - w^*}  \leq  \frac{4t\gamma^{2^{t-t_0}}}{\kappa (1 - \gamma^{2^{t-t_0}})},
		\end{align*}
		and  its communication complexity is
		\begin{align}
			T = O\BigP{\log \log \frac{1}{\epsilon}}.  \label{E:Th31}
		\end{align}
	\end{theorem}
{The proof can be found in \ref{Proof:Th:3}.} We see that controlling $\delta$ gives a trade-off between computation and communication complexities, where very small $\delta$ increases the computation iterations but may significantly reduce the communication cost with quadratic convergence, and vice versa. 
\subsection{Comparison with other distributed methods}
If  a first-order method such as GD is used for federated edge learning, the edge aggregator first receives $\nabla f_i(w_t), \forall i$,  updates
\begin{align}
	w_{t+1} = w_{t} - \eta \biggP{\frac{1}{n}\SumLim{i=1}{n} \nabla f_i(w_t)}, \label{E:GD}
\end{align}
and then sends $w_{t+1}$ to all workers for next iteration. By choosing $\eta = \frac{2}{\lambda + L}$, it has been shown that the communication complexity of GD for $\epsilon$-optimal solution \cite[Theorem 2.1.5]{nesterovIntroductoryLecturesConvex2013b} is $\kappa \log \frac{1}{\epsilon}$.
We note that this distributed GD \eqref{E:GD}  is a common approach in data center-type distributed machine learning \cite{liScalingDistributedMachine2014g,liCommunicationEfficientDistributed2014}, which is different from FedAvg-type GD in \cite{mcmahanCommunicationEfficientLearningDeep2017i,wangAdaptiveFederatedLearning2019b}. While FedAvg-type GD (or SGD) allows each device/worker to update its local model multiple times using local gradient information (thus its alternative name is Local GD/SGD \cite{stichLocalSGDConverges2018b}), the update in \eqref{E:GD} requires each client/worker to calculate its gradient once and send to the aggregator. Compared to first-order methods, without multiple local model updates, \OurAlg is more in line with distributed GD in \eqref{E:GD}  than FedAvg-type GD algorithms. 

Regarding other related Newton-type methods,  DiSCO \cite{zhangDiSCODistributedOptimization2015a} requires several communication rounds between the  aggregator and workers even for one Newton direction update, whereas GIANT \cite{wangGIANTGloballyImproved2018b} requires homogenous data, in which data must be collected centrally, shuffed, and distributed evenly to each node. Thus, both DiSCO and GIANT are only applicable to data center-type distributed machine learning. On the other hand, DANE \cite{shamirCommunicationefficientDistributedOptimization2014a} and FEDL \cite{dinhFederatedLearningWireless2020a}, approximate Newton-type methods,  are comparable to \OurAlg have  been shown to be a good candidate for federated learning. In DANE and FEDL, two communication rounds are used in one global iteration. In the first round, the workers compute the local gradients $\nabla f_i(w_t)$, which are then aggregated into the global gradient $\nabla f(w_t)$. In the second round, each worker solves a special local optimization using $f_i(w)$ and $\nabla f(w_t)$. It has been shown that to achieve an $\epsilon$-accurate solution,\footnote{The definition of $\epsilon$-accuracy in \cite{shamirCommunicationefficientDistributedOptimization2014a,dinhFederatedLearningWireless2020a} is $f(w_t) - f(w^*) \leq \epsilon$.} DANE requires ${O}\BigP{\frac{\kappa^2}{D_i} \log (d n)\log \frac{1}{\epsilon}}$ iterations while FEDL requires $O\BigP{\frac{1}{\Theta}\log \frac{1}{\epsilon}}$ with $\Theta \in (0,1)$. 
A comparison of communication and computation complexities of \OurAlg, distributed GD, FEDL, and DANE for a ridge regression task (quadratic loss) is summarized in Table \ref{T:compare}.

\section{Experimental Results and Discussion} \label{Sec:Sim}
\begin{figure}[t!]
	\centering
	\includegraphics[width=0.9\linewidth]{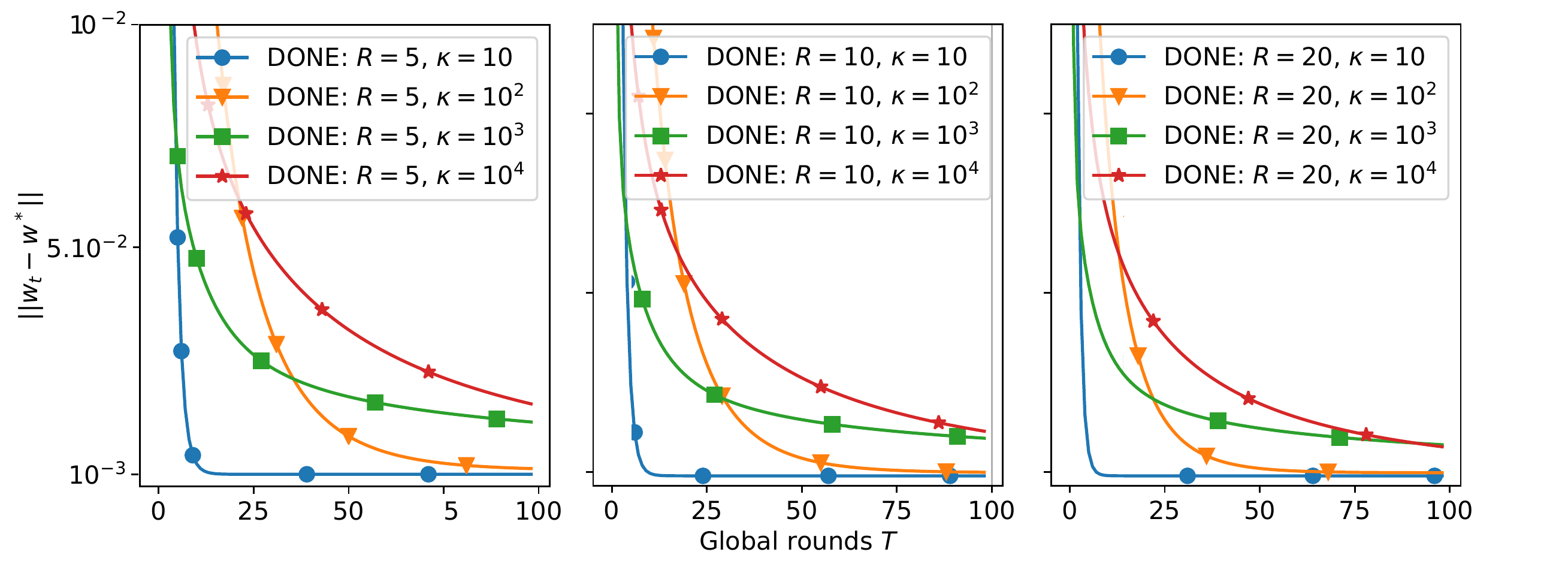}
	\vspace{-3mm}
	\caption{Effects of $\kappa$ on convergence of \OurAlg ($\alpha = 0.05, R = 5, 10, 20$) on regression task. Larger $\kappa$ slows down the convergence of \OurAlg; therefore, \OurAlg requires more local update $R$ to reduce the approximate error $\delta$.}
	\label{F:Synthetic_kappa}
	\vspace{-3mm}
\end{figure}
In this section, we evaluate the performance of \OurAlg when the data sources across edge workers are non-i.i.d. and heterogeneous. We first show the effect of various hyperparameters and how to choose those parameters for different scenariosc. We then compare \OurAlg  with the Newton's method \eqref{E:Newton}, FEDL, DANE, GIANT, and GD \eqref{E:GD} using real datasets. 
The experimental outcomes show that \OurAlg has the similar performance to the Newton's method with the same hyperparameters settings and achieves the performance improvement over FEDL, DANE, and GD in terms of testing accuracy, running time, and communication cost when all algorithms have a same target accuracy.
\begin{figure*}[t!]
	\centering
	{\includegraphics[scale=0.26]{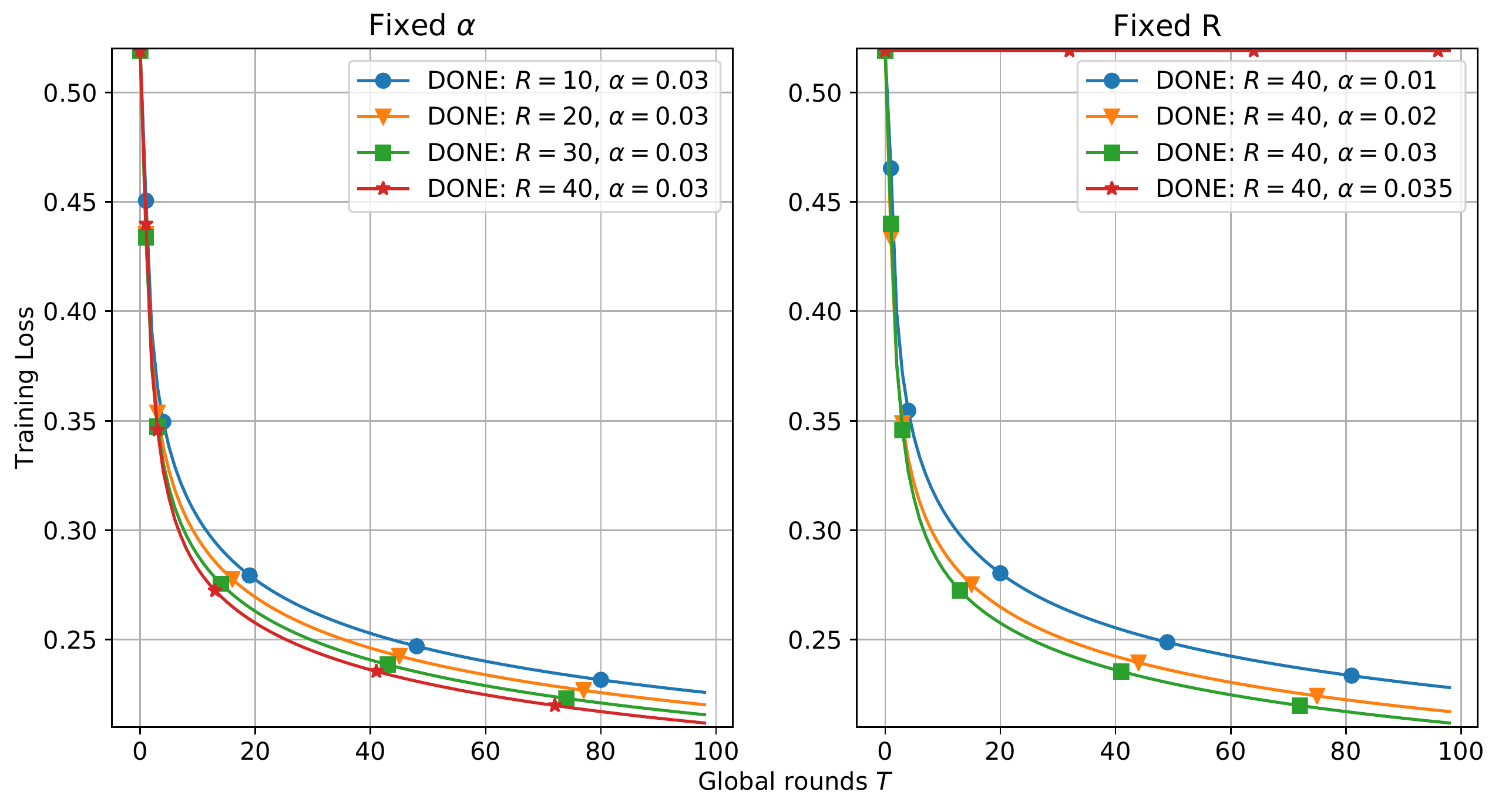}}\quad
	{\includegraphics[scale=0.26]{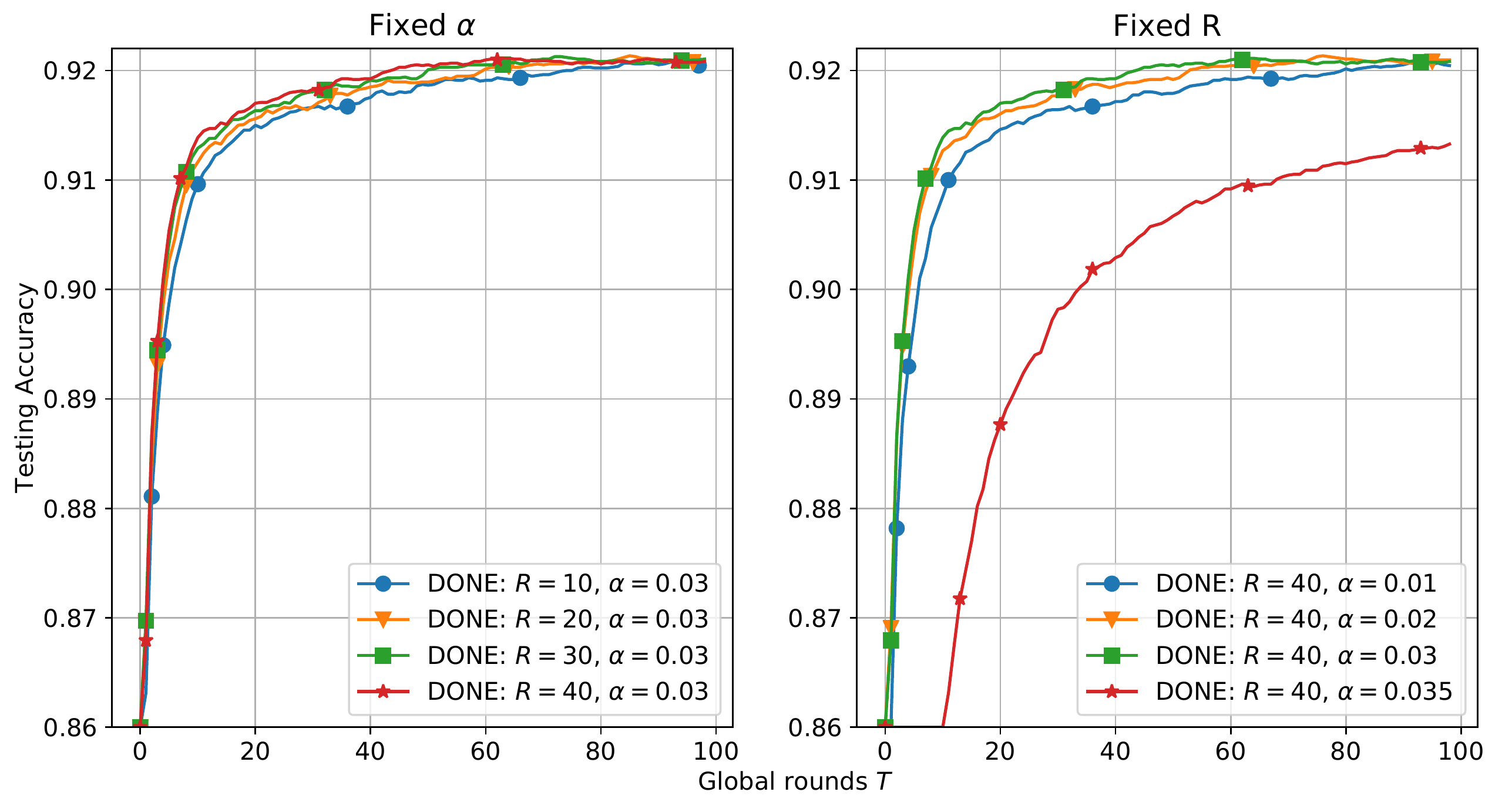}}
	\vspace{-2mm}
	\caption{Effects of various values of $\alpha$ and $R$  on MNIST.}
	\vspace{-3mm}
	\label{F:Mnist_R_alpha}	
\end{figure*}
\begin{figure*}[t!]
	\centering\
	{\includegraphics[scale=0.26]{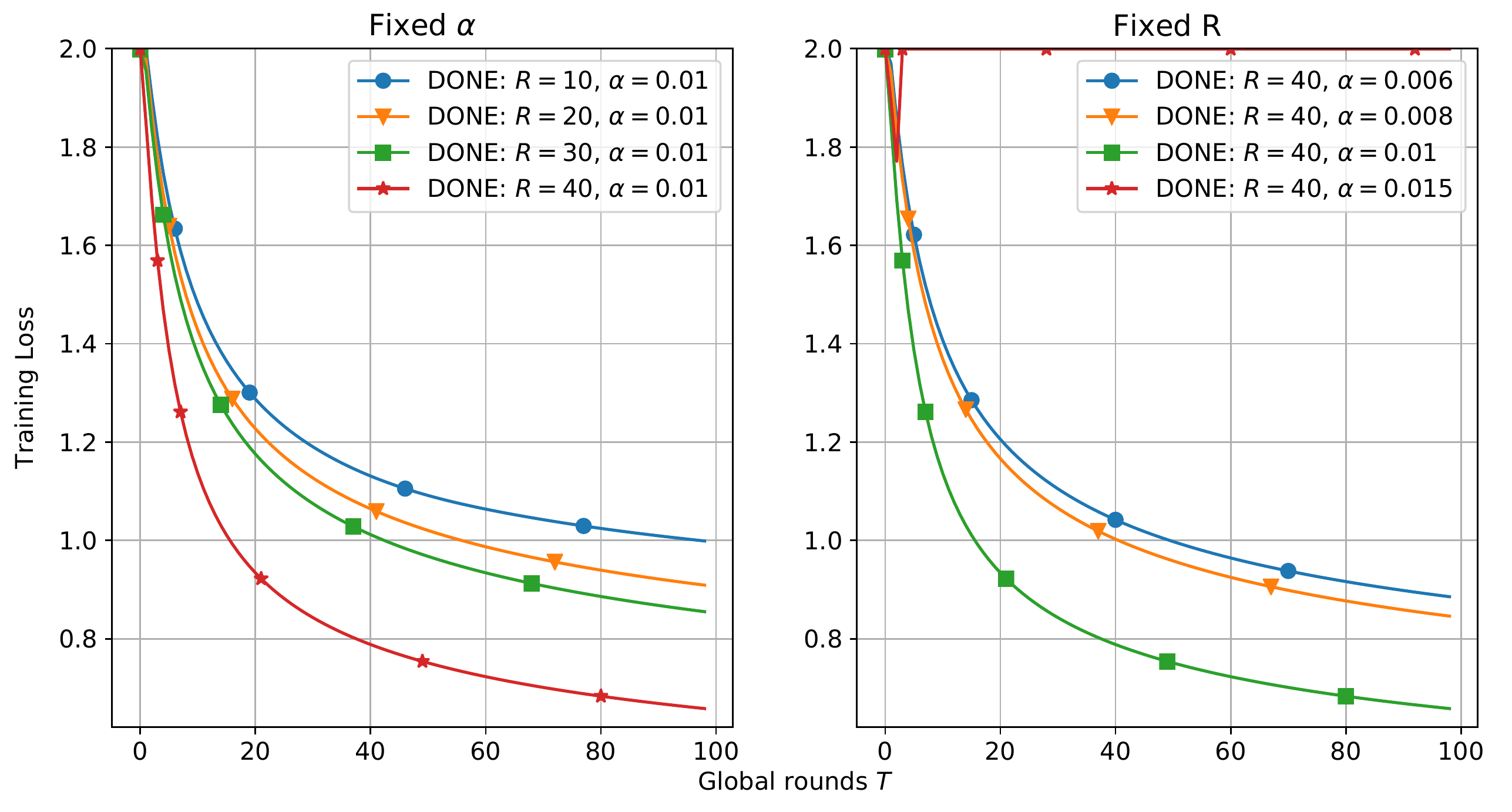}}\quad
	{\includegraphics[scale=0.26]{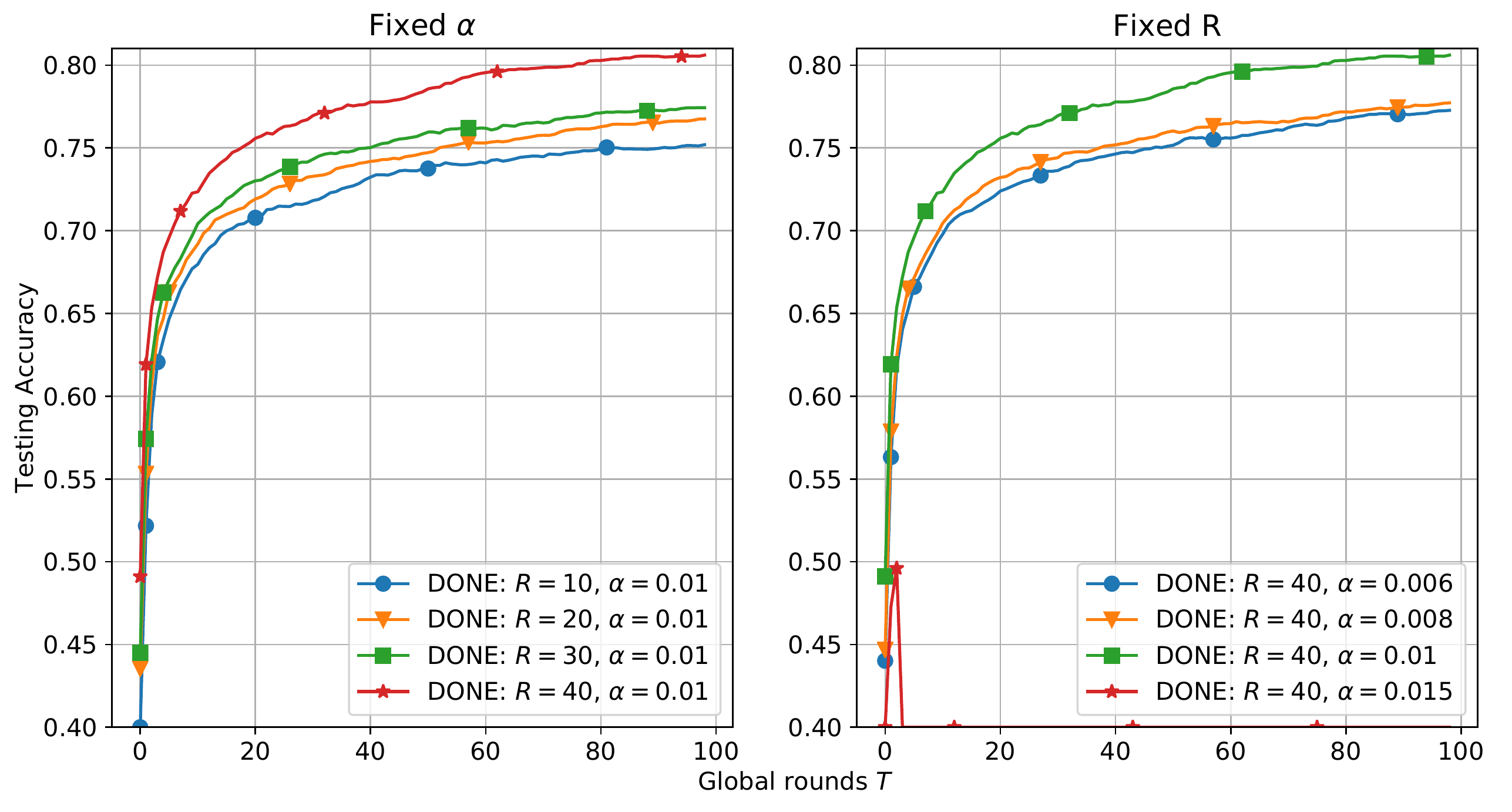}}
	\vspace{-2mm}
	\caption{Effects of various values of $\alpha$ and $R$  on FEMNIST.}
	\vspace{-3mm}
	\label{F:Nist_R_alpha}	
\end{figure*}
\begin{figure*}[t!]
	\centering
	{\includegraphics[scale=0.26]{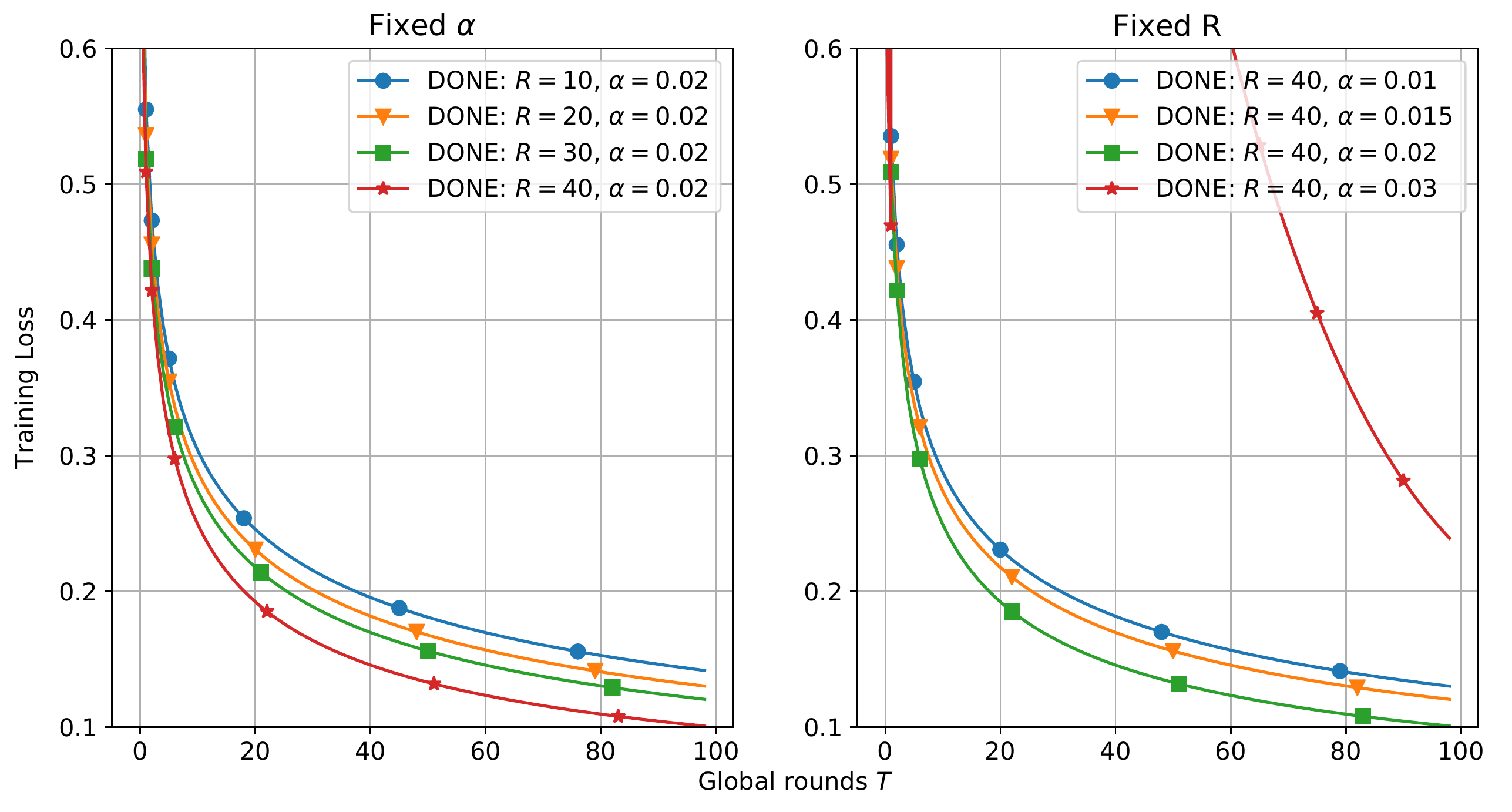}}\quad
	{\includegraphics[scale=0.26]{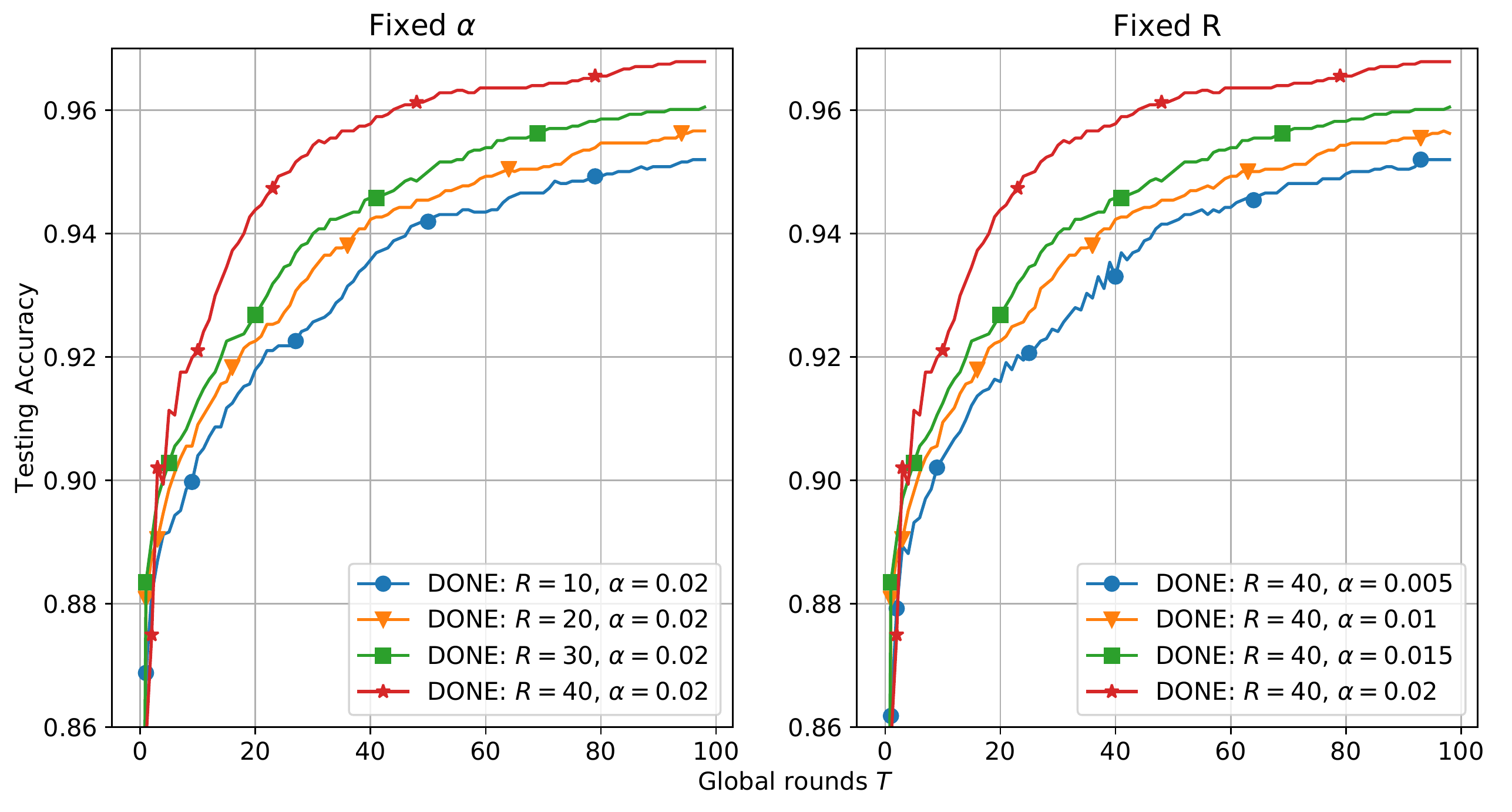}}
	\vspace{-2mm}
	\caption{Effects of various values of $\alpha$ and $R$  on Human Activity.}
	\vspace{-3mm}
	\label{F:Humnan_R_alpha}	
\end{figure*}
\begin{figure}[t!]
	\centering
	\begin{subfigure}{.8\columnwidth}
		\includegraphics[width=\columnwidth]{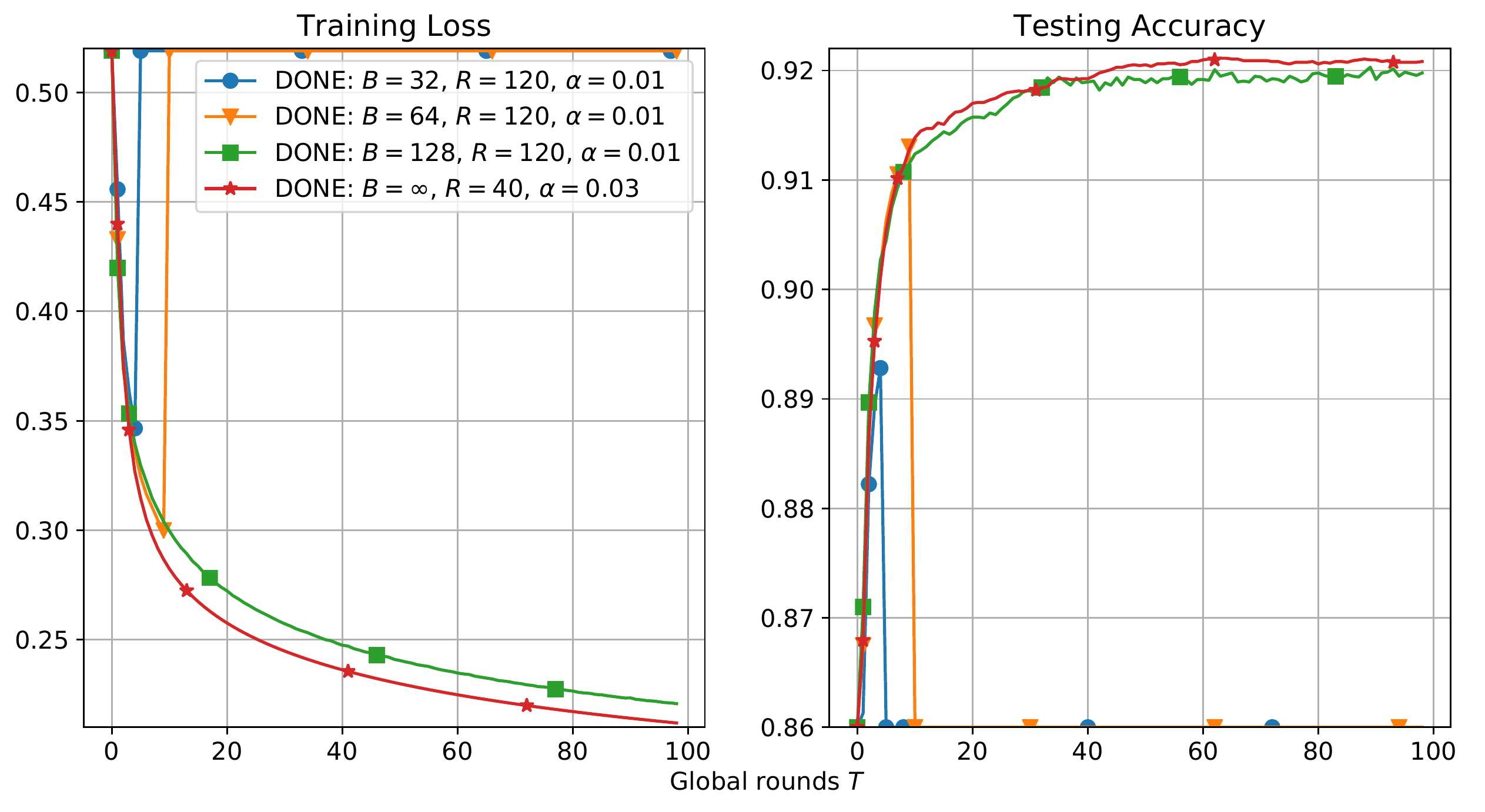}
		\vspace{-5mm}
		\caption{MNIST}
		\label{F:Mnist_batch}	
	\end{subfigure}
	
	\begin{subfigure}{.8\columnwidth}
		\includegraphics[width=\columnwidth]{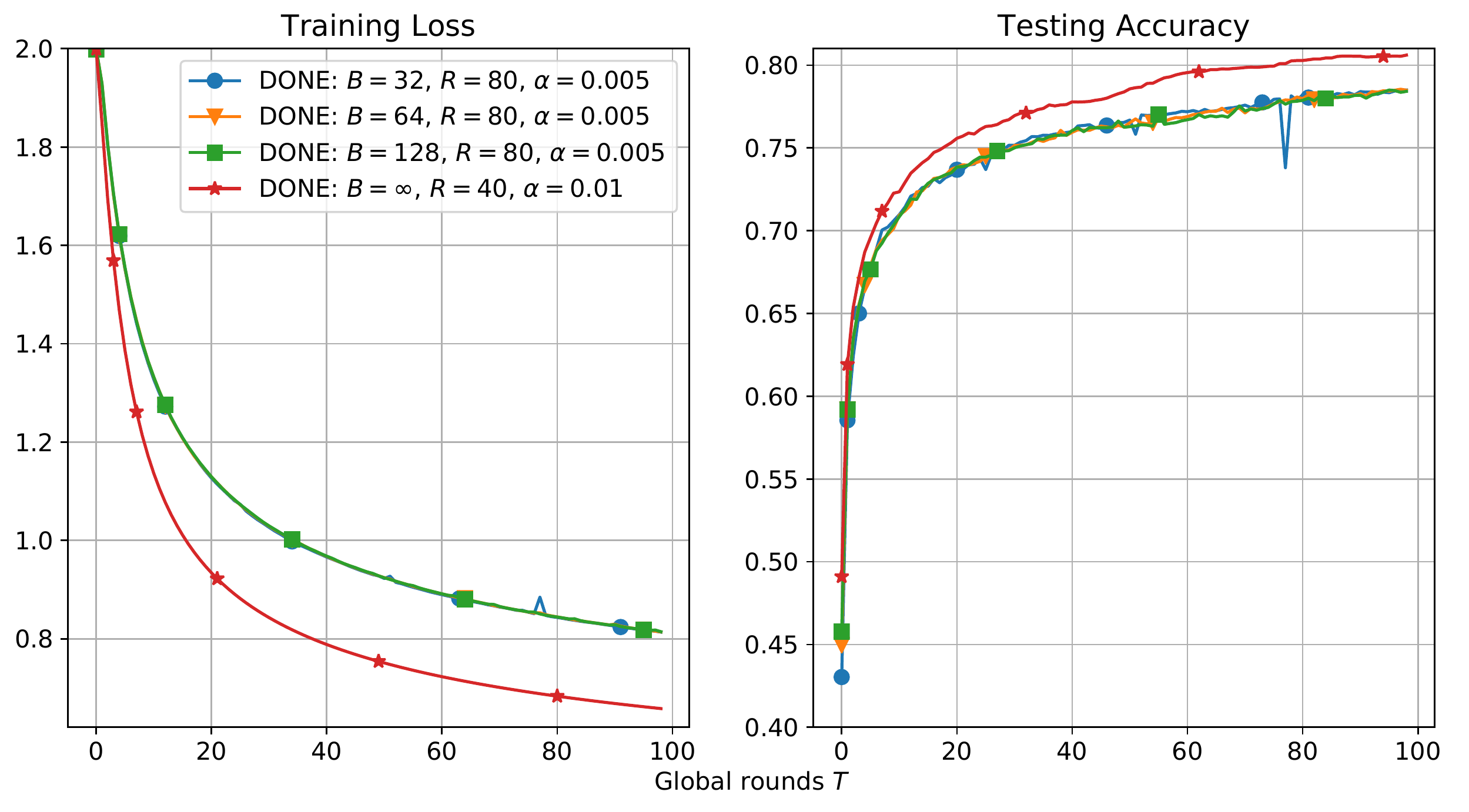}
		\vspace{-5mm}
		\caption{FEMNIST}
		\label{F:Nist_batch}	
	\end{subfigure}
	\begin{subfigure}{.8\columnwidth}
		\includegraphics[width=\columnwidth]{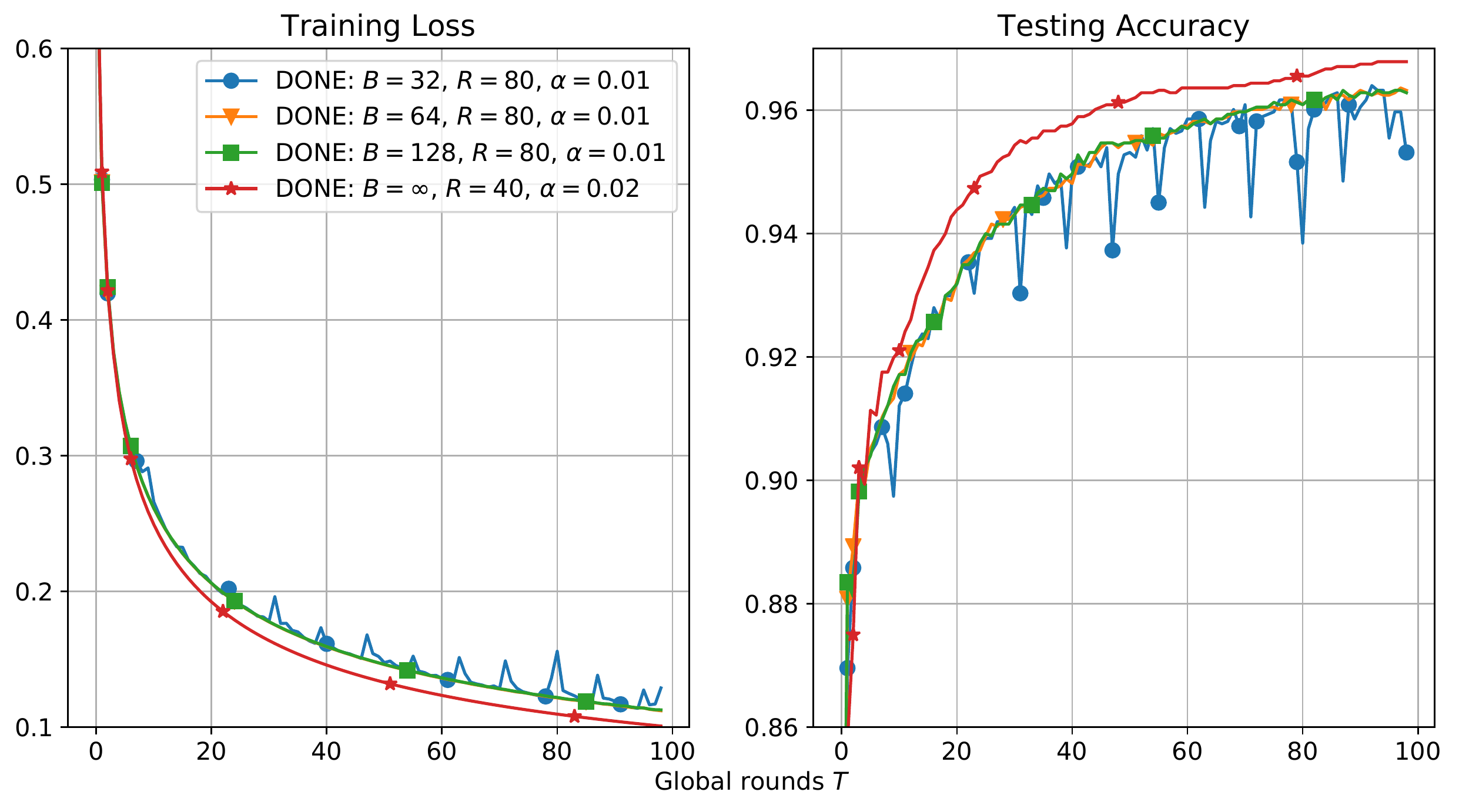}
		\vspace{-5mm}
		\caption{Human Activity.}
		\label{F:Humnan_batch}	
	\end{subfigure}\	\vspace{-2mm}
	\caption{Effects of mini-batch sampling. \OurAlg are more stable with the larger batch size.}
\end{figure}

\begin{figure}[t!]
	\centering
	\begin{subfigure}{.8\columnwidth}
		\includegraphics[width=\columnwidth]{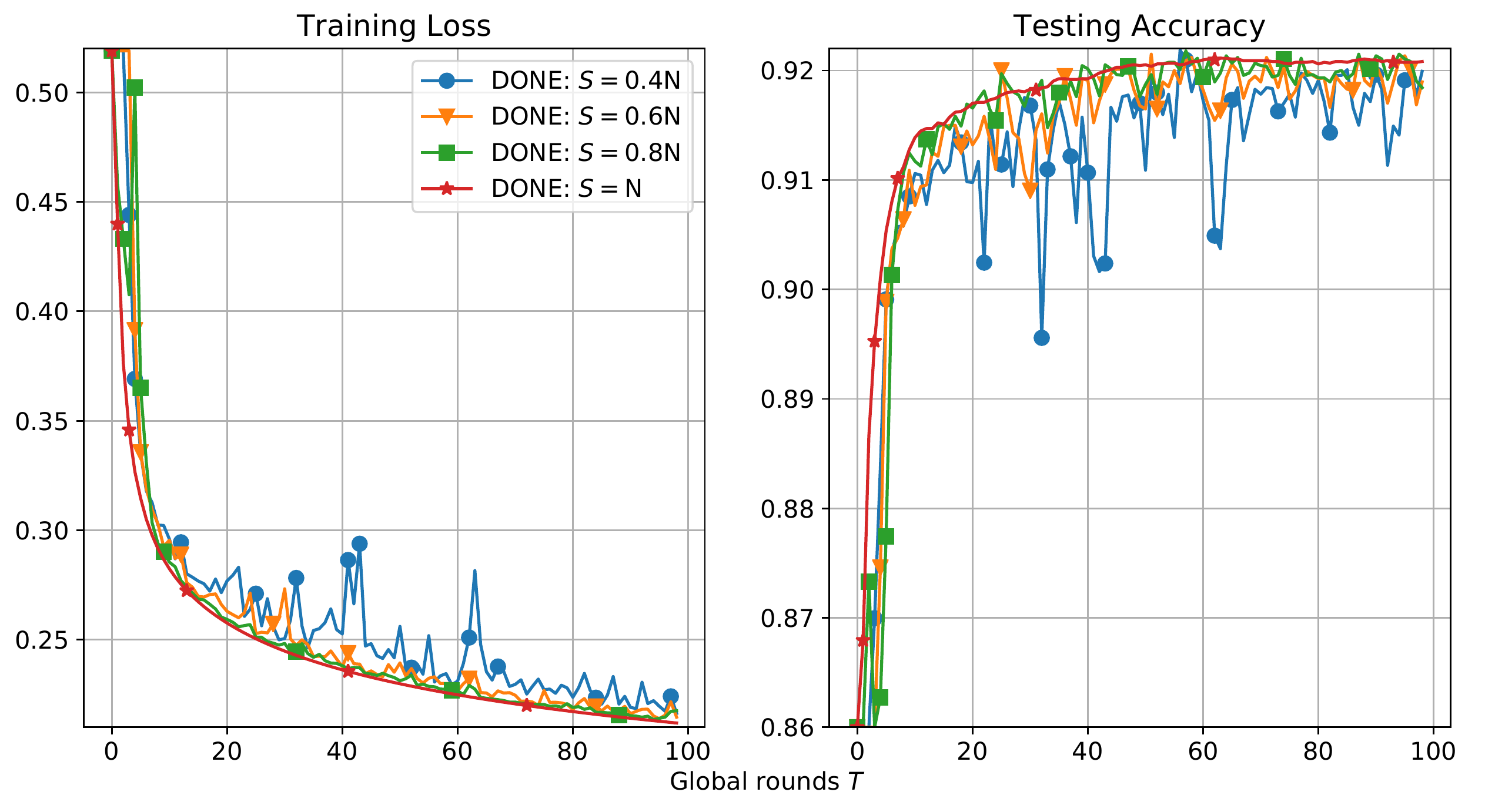}
		\vspace{-5mm}
		\caption{MNIST}
		\label{F:Mnist_edge}	
	\end{subfigure}
	
	\begin{subfigure}{.8\columnwidth}
		\includegraphics[width=\columnwidth]{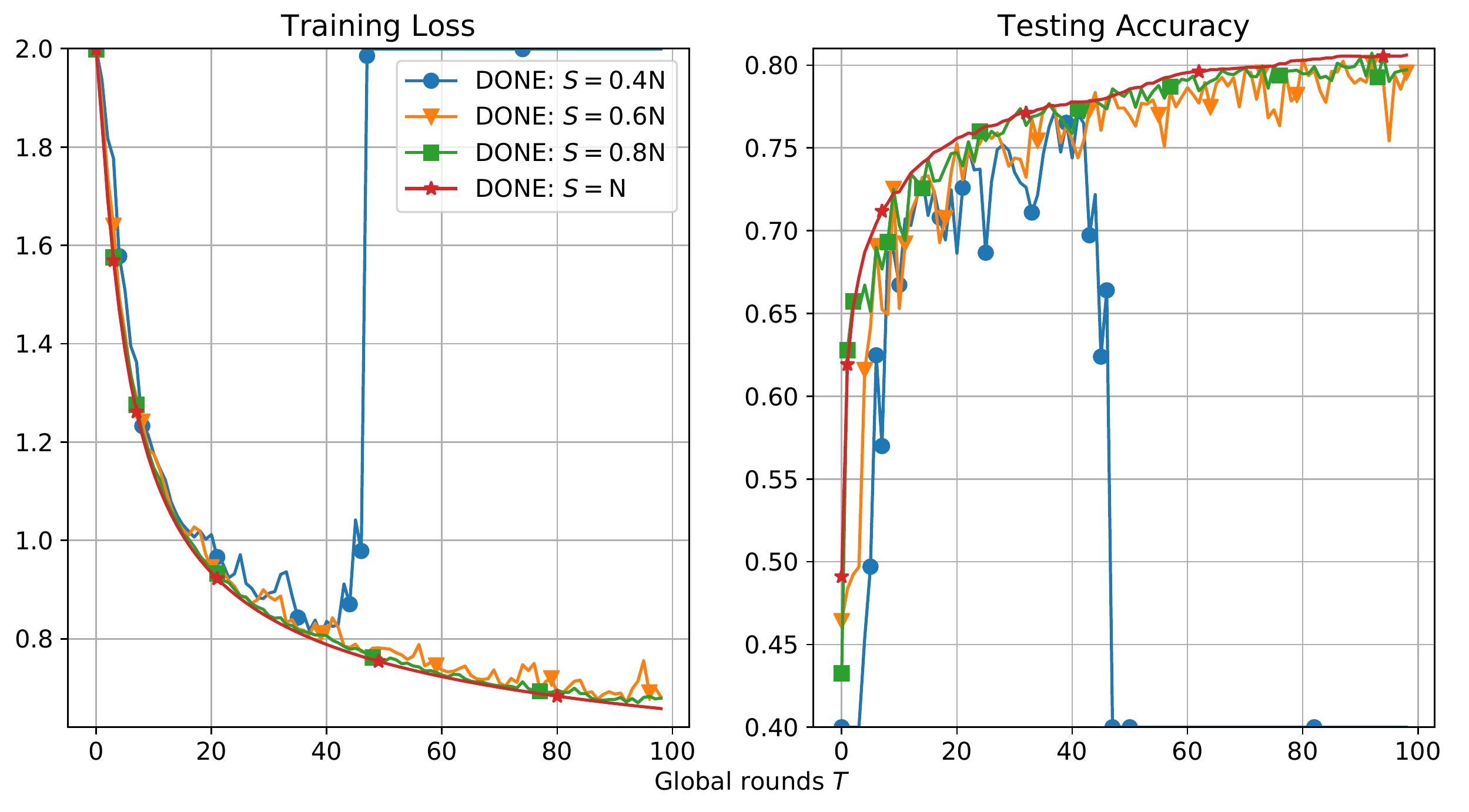}
		\vspace{-5mm}
		\caption{FEMNIST}
		\label{F:Nist_edge}	
	\end{subfigure}
	\begin{subfigure}{.8\columnwidth}
		\includegraphics[width=\columnwidth]{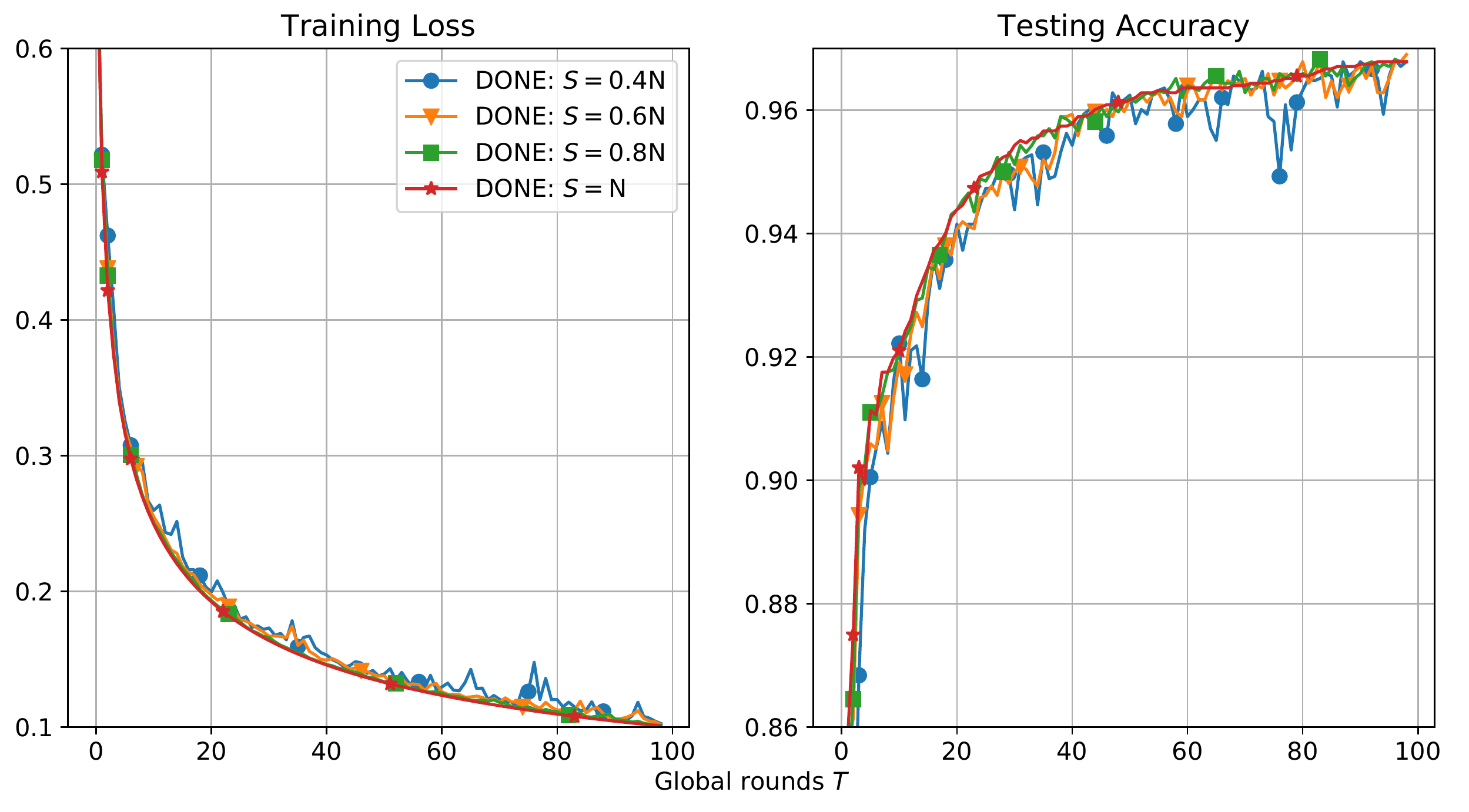}
		\vspace{-5mm}
		\caption{Human Activity.}
		\label{F:Humnan_edge}	
	\end{subfigure}
	\vspace{-2mm}
	\caption{Effect of sampling a subset of edge workers.}
\end{figure}
\subsection{Experimental Settings}

We implement several classification and regression experiments. For  classification tasks, to compare the performance of \OurAlg with other methods, we use three real datasets including MNIST, FEMNIST, and Human Activity Recognition generated in federated settings. We also use a synthetic dataset for a linear regression task, to monitor the effect of the condition number $\kappa$. We distribute the complete data to $n = 32$ edge workers ($n = 30$ only for Human Activity) and randomly split the data into two parts: 75\%  for training and 25\% for validation. The detail of all datasets are provided as follow: 

\textbf{Synthetic}: To generate non-i.i.d. data, each egde worker $i$ has a collection of $D_i$  samples  $\{a_j, y_j\}_{j=1}^{D_i}$  following the linear regression model $y_j = \innProd{w^*, a_j} + c_j$ with $w^*$, $a_j \in \mathbb{R}^d$, $c_j \in \mathbb{R}$, $a_j \thicksim \mathcal{N}(0,\sigma_j\Sigma)$, where $\sigma_j \thicksim \mathcal{U}(1, 30)$, $c_j \thicksim \mathcal{N}(0, 1)$, and $\Sigma \in \mathbb{R}^{d \times d}$ is a diagonal covariance matrix with $\Sigma_{i,i} = i^{-\tau}, i = 1,\ldots, d$. 
The main purpose of synthetic data is to control the condition number $\kappa$. By setting $\tau = \frac{\log(\kappa)}{\log(d)}$, $\kappa = d^\tau$ is the ratio between the maximum and minimum covariance values of $\Sigma$. To model a heterogeneous setting, each edge has a different data size in the range $[540, 5630]$. 

\textbf{MNIST} \cite{lecunGradientbasedLearningApplied1998a}: A handwritten digit dataset including 70,000 samples and 10 labels. In order to simulate a heterogeneous and non-i.i.d. data setting, each worker has only 3 labels and the data sizes vary in the range $[219, 3536]$.
	 
 \textbf{FEMNIST}: A dataset partitioned from Extended MNIST \cite{cohenEMNISTExtensionMNIST2017} which includes  62-class digit  following \cite{liFederatedOptimizationHeterogeneous2020a}
 . To generate federated setting, only 10 lower case characters (‘a’-‘j’) are selected and distributed 32 egdes (5 classes per edge).
	  
 \textbf{Human Activity Recognition} \cite{anguitaPublicDomainDataset2013}: A dataset collected from mobile phone accelerometers and gyroscope of 30 individuals, each performing one of six different activities: sitting, walking, walking upstairs, walking downstairs, lying down and standing. This dataset naturally captures the federated non-i.i.d. and heterogeneous characteristics. By considering each individual as an edge, we have 30 edges in total and the data size of each edge is in the range $[281, 383]$.

In our experiment, the regression task on the synthetic dataset uses a linear  regression model with mean squared error loss, and the classification task on real datasets uses multinomial logistic regression (MLR) models with cross-entropy loss.
We implement all algorithms using PyTorch version 1.8.0 {and evaluate on Tesla K80 GPU}. Each experiment is run 10 times for statistical reports.
Code and datasets are available online\footnote{https://github.com/dual-grp/DONE}.

\subsection{Effect of $\kappa$}
We first verify our theoretical findings by  observing the convergence behavior of \OurAlg on a wide range values for $\kappa$, including small ($\kappa = 10$), medium ($\kappa \in \{10^2, 10^3\}$), and large ($\kappa = 10^4$) on the synthetic dataset in Fig~\ref{F:Synthetic_kappa}. Using  the linear regression model, we can obtain $\lambda_{i, max}$, the  largest eigenvalues of the Hessian of $f_i$ of each worker $i$. We then choose $\alpha \leq \frac{1}{ \hat \lambda_{max}}$, e.g., $\alpha = 0.05$ and $\alpha \leq \frac{1}{R}$ following Theorem \ref{Th:1}. It can be seen that using choosen values of $\alpha$ and $R$ allows DONE to converge approximately to the solution with all settings of $\kappa$. However, with given the same error tolerance, larger $\kappa$ requires larger $R$ to reduce the approximation error $\delta$, which is verified in Lemma~\ref{Lem:2}.
\begin{table}[t!]
	\small
	\centering
	\caption{ Performance comparison on three real datasets. We fix $R = 40, T = 100$ and for all algorithms. $\gamma$ is the learning rate of DANE, FEDL, and GD.}
	\vspace{-1mm}
			\setlength\tabcolsep{4pt} 
	\small
	\begin{tabular}{l|l|lll}
		\multicolumn{1}{l|}{\begin{tabular}[c]{@{}c@{}} Dataset \end{tabular}} & Algorithm & $\alpha$ ($\gamma)$  & Accuracy &  {Running Time  (ms)}\\ \hline
		\multirow{4}{*}{MNIST}      
		& DONE      &     $0.03$               &  $\textbf{92.11}    \pm 0.01$       &      {$ 399.00   \pm  2.52$  }    \\
		& FEDL      &       $ 0.04   $          &          $ 91.89      \pm 0.01 $         &   {$ 660.27        \pm  2.87$   }      \\ 
		& DANE      &       $ 0.04   $          &          $ 91.84      \pm 0.01 $       &    {$742.22         \pm   3.11$ }        \\ 
		& GIANT      &                &          $ 91.89      \pm 0.01 $            &  {$464.52     \pm  2.41$    }    \\ 
		& Newton    &       $0.03 $            &      $ \textbf{92.11}      \pm 0.01  $      &  {$ 1176.60     \pm  3.57 $       }            \\
		& GD        &            $0.2 $        		&  $91.35           		\pm 0.02$           &    {$42.79    \pm  3.09 $}       \\ \hline
		\multirow{4}{*}{\begin{tabular}[c]{@{}c@{}} FE-\\MNIST \end{tabular}}      
		& DONE      &        $ 0.01 $           &     $\textbf{80.60}   \pm 0.02 $                  &   {$372.97    \pm  2.16$   }     \\
		& FEDL      &            $ 0.01 $       &    $  78.28          \pm 0.01 $               & {$486.08   \pm  2.11 $ }      \\
		& DANE      &       $ 0.01   $          &          $ 77.57       \pm 0.01 $            &   {$490.28     \pm  2.25$}      \\
		& GIANT      &                &          $ 79.61      \pm 0.01 $         &   {$424.25     \pm    1.16$}   \\ 
		& Newton    &       $ 0.01$             &      $\textbf{80.60}   \pm 0.02 $          &  {$911.26        \pm 3.16 $  }        \\
		& GD        &      $0.02      $         &           $60.58     \pm 0.03 $                   &  {$39.60    \pm 2.68$  } \\ \hline
		\multirow{4}{*}{\begin{tabular}[c]{@{}c@{}} Human\\ Activity\end{tabular}}       
		& DONE      &       $0.02     $         &          $96.78    \pm 0.01 $         & {$216.42      \pm 1.35 $     }        \\
		& FEDL      &    $   0.05  $            &    $     95.90      \pm 0.01$         & {$ 246.13       \pm 1.49 $    }           \\
		& DANE      &       $ 0.05   $          &          $95.82     \pm 0.01 $       & {$256.79     \pm 1.53 $    }        \\
		& GIANT      &                &          $ 96.13     \pm 0.02 $             & {$354.33    \pm  1.67$ }    \\ 
		& Newton    &      $ 0.02     $        &            $96.78 \pm 0.01$         & {$583.83      \pm  2.71$  }         \\
		& Newton    &      $ 0.03     $        &        $\textbf{96.90} \pm 0.01$  &             \\
		& GD        &     $ 0.1       $        &  $ 80.02  \pm 0.02 $               &   {$23.84     \pm  1.95$   }    \\ \hline
	\end{tabular}
	\label{T:compare_performance}	
		\vspace{-3mm}
\end{table}

\begin{table}[t!]
	\small
		\centering
			\setlength\tabcolsep{2pt} 
	\caption{Communication times $T$ for all algorithms to achieve the same target accuracy.}
	\begin{tabular}{l|llll|l}
	\multicolumn{1}{l|}{\multirow{2}{*}{Dataset}} & \multicolumn{4}{c}{Algorithm} & \multicolumn{1}{|c}{\multirow{2}{*}{\begin{tabular}[c]{@{}l@{}}Target\\ Accuracy\end{tabular}}}\\
		\multicolumn{1}{l|}{}                    & DONE  & GIANT & FEDL  & DANE    &\multicolumn{1}{c}{}   	 \\ \hline
		MNIST                                        &  \textbf{28}     &     59   &  70      &  100   & \textbf{91.84}  \\ \hline
		FEMNIST                                      &      \textbf{31} &  62     &   75    &   100 &  \textbf{77.57}  \\ \hline
		Human Activity                               &     \textbf{55}  &   77    & 86      & 100  &\textbf{95.82 }  \\ \hline
	\end{tabular}
	\label{T:time}	
		\vspace{-3mm}
\end{table}
\subsection{Effect of Hyper-parameters: $\alpha$, $R $}
We show the impact of wide ranges of $\alpha$, $R $ on the convergence of \OurAlg for MNIST, FEMNIST, and Human Activity in Figs.~\ref{F:Mnist_R_alpha}, \ref{F:Nist_R_alpha}, and \ref{F:Humnan_R_alpha}, respectively. To monitor the effect of $\alpha$, we ﬁx the value of  $R $ and vice versa. The results demonstrate that there exist sets of sufficiently small $\alpha$ and large $R$ following the condition such that $\alpha < \frac{1}{R}$ allowing \OurAlg to converge. We observe that both larger $\alpha$ and $R$ speed up the convergence of \OurAlg as the larger $R$ allows \OurAlg to approximate true Newton direction closely. However, increasing $R$ comes at a cost of higher local computation, and increasing $\alpha$ ($\alpha \geq 0.035$ for MNIST, $\alpha \geq 0.015$ for FEMNIST, and $\alpha \geq 0.03$ for Human Activity) can lead to divergence of \OurAlg.  Both $R$ and $\alpha$  should be tuned carefully depends on the heterogeneity of data. The more heterogeneous data is, the less value of $\alpha$ and the higher value of $R$ is considered to reduce the approximate term $\delta$. In federated edge computing where each edge worker has a powerful computational capacity, it is reasonable to use larger $R$ and smaller $\alpha$ to handle the heterogeneity and also to reduce the cost of communication.
\subsection{Effect of Mini-batch Sampling}
Even though our analysis of \OurAlg requires computing full-batch Hessian-gradient products, we also consider an additional case where we sample a mini-batch of size $B$ (where $B \in \set{32, 64, 128}$) in each round to approximate the true Hessian-gradient product. From the experimental results, using a mini-batch requires a smaller value of $\alpha$  than that of full-batch. In Figs.~\ref{F:Mnist_batch}, \ref{F:Nist_batch}, and \ref{F:Humnan_batch}, by reducing the value of $\alpha$ and increasing the value of $R$ correspondingly, the performance of \OurAlg using mini-batches is close to that of \OurAlg using full batches. In addition, using small mini-batches can lead to instability of \OurAlg: e.g., when $B=32$ \OurAlg diverges in case of MNIST and becomes less stable in the cases of FEMNIST and Human Activity.
\subsection{Effect of Edge Worker Sampling}
In practice, besides using mini-batches to reduce computation, it is critical to address the straggler's effect. We consider a scenario when a subset of workers of size $S$  is selected randomly for aggregation. We keep using the same experimental setting as above but randomly select the value of  $S$ in $\{N, 0.8N, 0.6N, 0.4N\}$. In Figs.~\ref{F:Mnist_edge}, \ref{F:Nist_edge}, and \ref{F:Humnan_edge}, \OurAlg converges in all choices of $S \geq 0.6N$. As expected,  larger $S$ allows \OurAlg to converge faster and be more stable. On the other hand, when $S \leq 0.4N$  we observe the deterioration of \OurAlg's performance, especially in the case of FEMNIST.

\subsection{Performance comparison with  distributed algorithms}
\begin{figure}[t!]
	\centering
	\begin{subfigure}{.8\columnwidth}
		\includegraphics[width=\columnwidth]{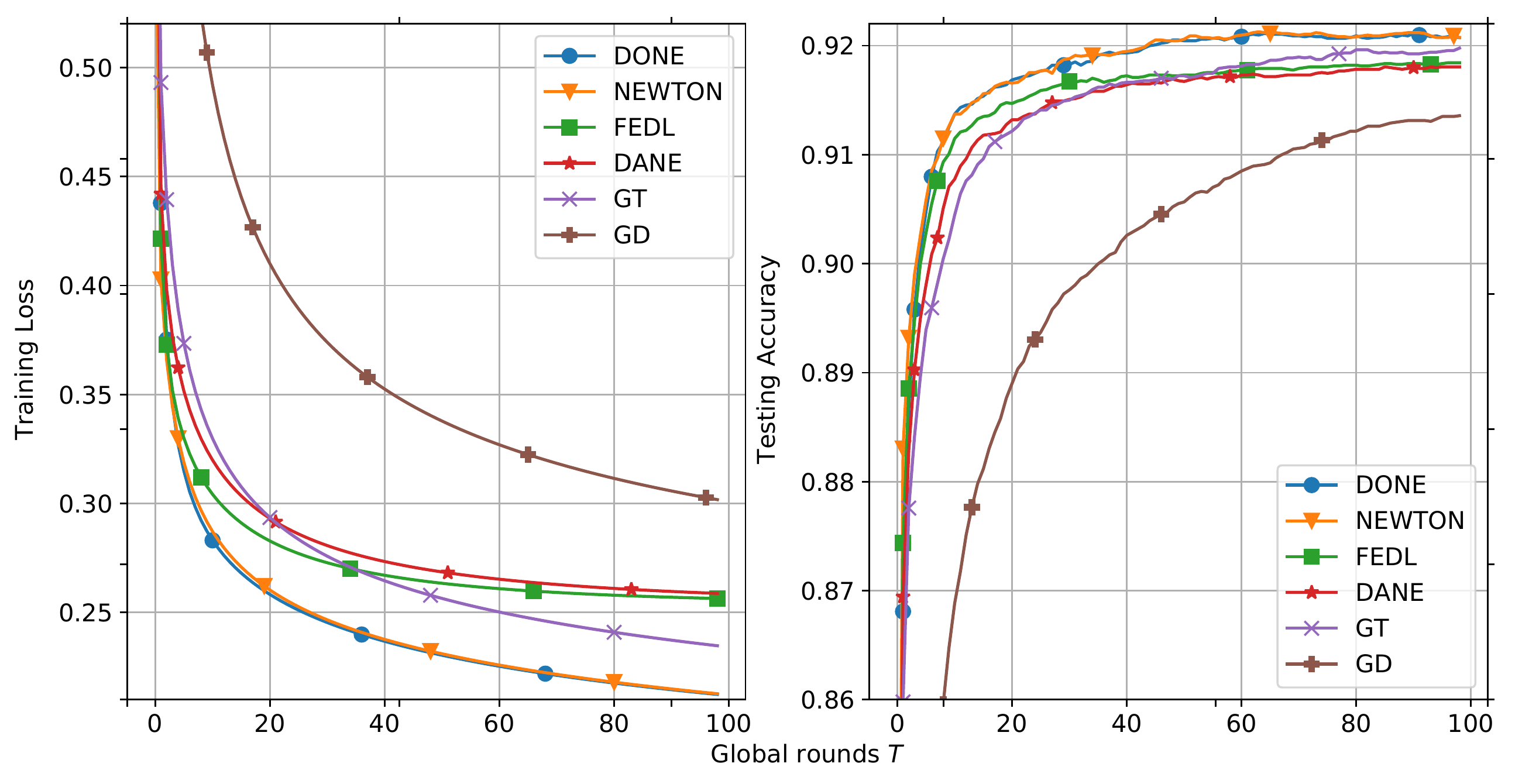}
		\vspace{-5mm}
		\caption{MNIST}
		\label{F:Mnist_al}	
	\end{subfigure}
	
	\begin{subfigure}{.8\columnwidth}
		\includegraphics[width=\columnwidth]{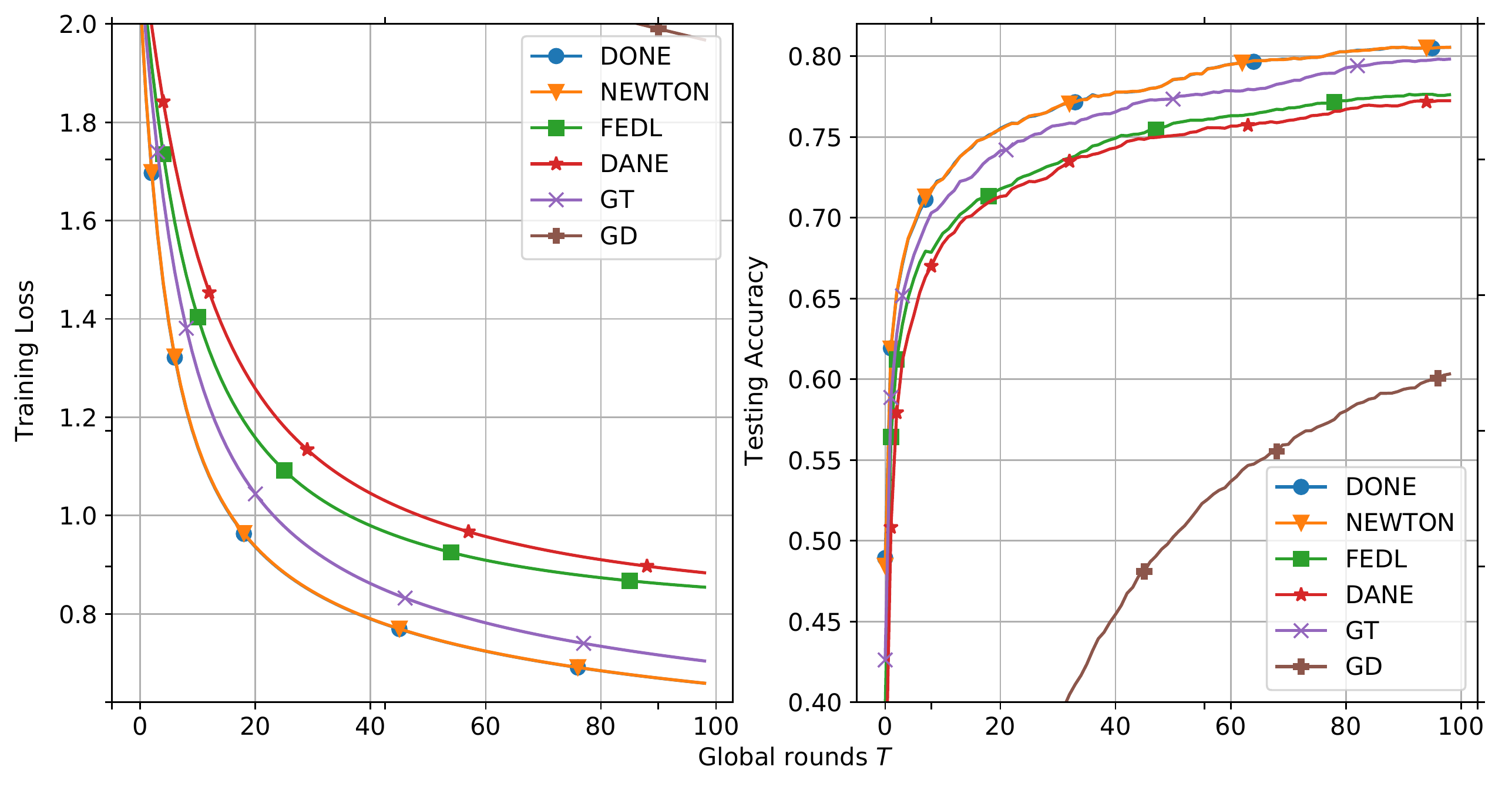}
		\vspace{-5mm}
		\caption{FEMNIST}
		\label{F:Nist_al}	
	\end{subfigure}
	\begin{subfigure}{.8\columnwidth}
		\includegraphics[width=\columnwidth]{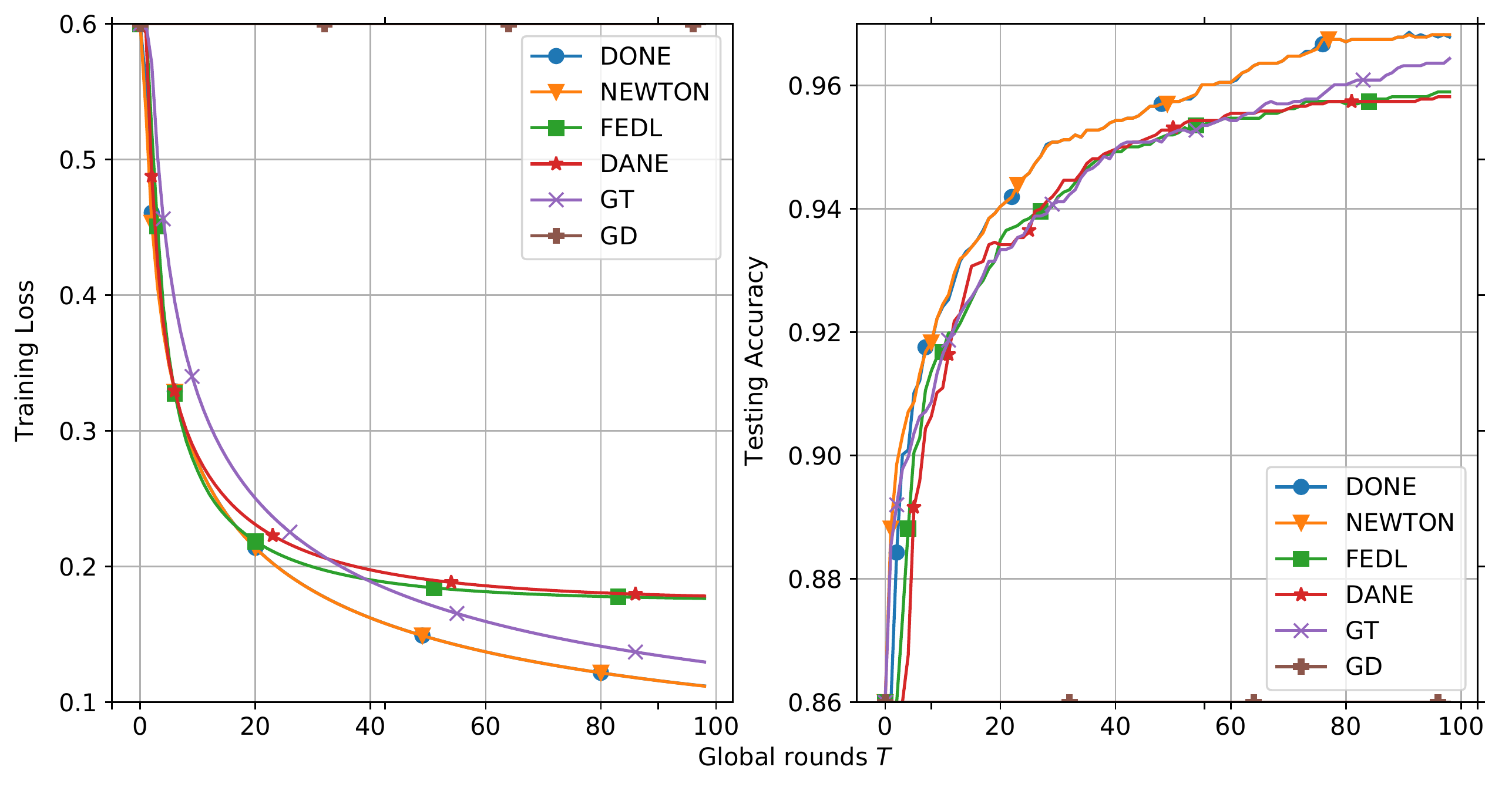}
		\vspace{-5mm}
		\caption{Human Activity.}
		\label{F:Humnan_al}	
	\end{subfigure}
	\vspace{-2mm}
	\label{F:convergence}	
	\caption{Convergence comparisons. We fix $R = 40, T = 100$ for all algorithms and use same values of $\alpha$ and $\gamma$ in Table \ref{T:compare_performance}.}
\end{figure}
We finally compare \OurAlg with the Newton's method, GD, DANE, and FEDL. For a fair comparison, we fix a same number of communication rounds ($T$) and the number of local updates ($R$) for all algorithms. We then {use grid search} to fine-tune the hyper-parameters w.r.t. the highest test accuracy and stability of each algorithm. 
 We fix $\eta  = 1$  for DANE and choose the best regularization parameter $\mu$ in $\{0, \lambda, 3\lambda\}$ for DANE and GIANT \footnote{We direct the reader to \cite{shamirCommunicationefficientDistributedOptimization2014a} for the meaning of DANE's parameters.}. We follow the setting of \cite{dinhFederatedLearningWireless2020a} for FEDL.
We also apply the Richardson iteration to the true Newton's method \eqref{E:Newton} since it is impractical to do the inverse Hessian. {Newton's method with Richardson iteration requires all edge workers to send the local Newton direction to the server for aggegation at each single local update, hence it actually takes $R.T$ communication rounds}. In comparison with Newton's method, two separate cases are considered: when the Newton's method has the same values of hyper-parameter $\alpha$ with DONE and uses fine-tuned $\alpha$. {We compare the accuracy and running time in Table \ref{T:compare_performance} and the convergence of all algorithms in Figs.~\ref{F:Mnist_al}, \ref{F:Nist_al}, and \ref{F:Humnan_al}}. The experimental results show that \OurAlg has a similar performance to Newton's method on the same set of $\alpha$ and $R$.  As highlighted in Theorem \ref{Th:1}, when $\alpha$ is small and $R$ is large enough, the Richardson iteration allows \OurAlg to catch up with the Newton's method. There is only a small gap (0.12\%) when Newton's method has a larger value of $\alpha$ which can be seen when using the Human Activity, a highly non-i.i.d dataset. To obtain similar performance with Newton's method in this scenario, {DONE needs to run more local iterations ($R = 50$)}. In the case of  FEMNIST and MNIST, \OurAlg and the Newton's method have the same value of fine-tuned  $\alpha$.  In comparison with others, \OurAlg improves from FEDL, DANE, GAINT, and GD in all real datasets. For MNIST, the improvement in test accuracy compared to FEDL, DANE, GIANT, and GD are approximately 0.22\%, 0.27\%, 0.22\%, and 0.76\%, respectively. The corresponding figures are 2.32\%, 3.03\%, 0.99\%, and 20.02\% for FEMNIST, and  0.88\%, 0.96\%, 0.65\%, and 16.76\% for Human Activity. Finally, to compare the communication and computation complexities among all algorithms, we first fix $T = 100$ and compare the running time of all algorithms. The results in Table \ref{T:compare_performance} show that DONE has the smallest running time. Additionally, in Table \ref{T:time}, we set the same target accuracy and compare the number of global rounds $T$ needed by each algorithm to achieve that accuracy. Overall, \OurAlg shows a marked improvement from GIANT, DANE, and FEDL, requiring much fewer iterations to achieve the same accuracy.
	

\section{Conclusion} \label{Sec:Conclusion}
In this work, we develop a distributed approximate Newton-type algorithm (\OurAlg) suitable for federated edge learning. We show that \OurAlg effectively approximates the true Newton direction using the Richardson iteration when the loss functions are strongly convex and smooth. Additionally, we specify that \OurAlg has global linear-quadratic convergence and provide its computation and communication complexity analysis. Finally, we experimentally verify our theoretical findings and demonstrate the competitiveness of our approach when compared with the distributed GD method, FEDL, and DANE, approximate distributed Newton-type algorithms.
\section{Proofs}
We provide proofs for the theorems and lemmas.
\subsection{Proof of Theorem \ref{Th:1}}
\label{Proof:Th:1}
By Richardson iteration convergence, it is straightforward to
have:  $\lim_{k \rightarrow \infty}x_k = A^{-1} b = x^*$ and
\begin{align}
\norm{x_k - x^*} \leq \norm{(I - \alpha A)^k} \norm{x_0 - x^*}. \label{E:richardson_distance}
\end{align}
Using the Richardson iteration, expanding $x_k$ recursively gives
\begin{align}
	x_k &= (I - \alpha A) x_{k-1} + \alpha b \nonumber \\
		&= (I - \alpha A)^k x_0 + \SumNoLim{j=0}{k-1}(I - \alpha A)^j \alpha b \label{E:x_k_expanded}.
\end{align}
Similarly, expanding $x_{i,k}$ recursively gives
\begin{align}
x_{i,k} = (I - \alpha A_i)^k x_{i, 0} + \SumNoLim{j=0}{k-1}(I - \alpha A_i)^j \alpha b \label{E:x_ik_expanded}.
\end{align}
Taking the average of all $x_{i,k}$ gives
\begin{align}
\frac{1}{n} \SumLim{i=1}{n} x_{i,k} &= \frac{1}{n} \SumLim{i=1}{n} \BigP{ \nbigP{I - \alpha A_i}^k x_{i,0} + \SumLim{j=0}{k-1}\bigP{I -  \alpha A_i}^j \alpha b} \nonumber\\
&= \frac{1}{n} \SumLim{i=1}{n} {  {\bigP{I -  \alpha A_i}^k x_{i,0} }} +\frac{1}{n} \SumLim{i=1}{n} { \SumLim{j=0}{k-1}\bigP{I -  \alpha A_i}^j \alpha b }. \label{E:x_ik}
\end{align}
Using the Taylor expansion on $(I -  \alpha A)^j$ gives
\begin{align}
\bigP{I-\alpha A}^j &= \binom{j}{0} I + \binom{j}{1}  (-\alpha A)^{1} + \binom{j}{2}  (-\alpha A)^{2} + ...  \nonumber\\
&= I -\alpha j A +  \alpha^2 \frac{j(j-1)}{2}A^2 + ...  \label{E:taylor}
\end{align}
Taking the sum of this term from $j=0$ to $k-1$ gives
\begin{align}
\SumLim{j=0}{k-1} \bigP{I-\alpha A}^j &= kI -\alpha  \frac{k(k-1)}{2} A  \nonumber\\ &+  \alpha^2 \frac{k(k-1)(k-2)}{3!}A^2 + ...  \label{E:binomial}
\end{align}
For any matrix $A$ and constant $\alpha$ satisfying $\norm{\alpha A_i} < 1, \forall i$, we can respectively express \eqref{E:taylor} and \eqref{E:binomial} as
\begin{align}
\bigP{I-\alpha A}^j &=  I -\alpha j A + O(\alpha^2 A^2)  \label{E:taylor2} \\
\SumLim{j=0}{k-1} \bigP{I-\alpha A}^j &= kI -\alpha  \frac{k(k-1)}{2} A + O(\alpha^2 k A^2) \label{E:binomial2}.
\end{align}
Substituting \eqref{E:taylor2} and \eqref{E:binomial2} into \eqref{E:x_k_expanded} gives
\begin{align}
x_k &= (I -\alpha k A + O(\alpha^2 A^2))x_0  \nonumber\\ 
&+ \BigP{kI - \frac{k(k-1)}{2} \alpha A + O(\alpha^2 k A^2)} \alpha b , \nonumber 
\end{align}
Similarly, substituting \eqref{E:taylor2} and \eqref{E:binomial2} into \eqref{E:x_ik_expanded} gives
\begin{align}
\frac{1}{n} \SumLim{i=1}{n} x_{i,k} &= 	\frac{1}{n}\SumLim{i=1}{n}\bigP{I -\alpha k A_i + O(\alpha^2 A_i^2)}x_{i,0} \nonumber \\
&+\frac{1}{n} \SumLim{i=1}{n} \bigP{kI - \alpha \frac{k(k-1)}{2} A_i}\alpha b  \nonumber \\
&+ \frac{1}{n} \SumLim{i=1}{n} O(\alpha^2 k A_i ^2)\alpha b.\nonumber
\end{align}
If $A = \frac{1}{n}\SumNoLim{i=1}{n} A_i$, $\nu =  \norm*{A^2 -	\frac{1}{n}\SumLim{i=1}{n} A_i^2}$, and $x_0 = x_{i,0}$ we can bound the distance between $x_k$ and $\frac{1}{n}\SumNoLim{i=1}{n}x_{i,k}$ as
\begin{align}
\norm*{x_k - \frac{1}{n}\SumLim{i=1}{n}x_{i,k}} & \leq  O\biggP{\alpha^2 \nu \norm*{x_0}} +   O\biggP{\alpha^3 k \nu \norm*{b}}.
\end{align}
Further, if $\alpha \leq \frac{1}{k}$, then
\begin{align}
\norm*{x_k - \frac{1}{n}\SumLim{i=1}{n}x_{i,k}} \leq
O\biggP{ \frac{\nu}{k^2} \bigP{\norm*{b} + \norm*{x_0}}}. \label{E:distributed_richardson_distance}
\end{align}
We have
\begin{align}
	 \norm*{\frac{1}{n}\SumNoLim{i=1}{n}x_{i,k} - x^*} &\leq \norm*{\frac{1}{n}\SumNoLim{i=1}{n}x_{i,k} - x_k} + \norm*{x_k - x^*} \label{E:triangle} \\
	& \leq O\biggP{ \frac{\nu}{k^2}  \bigP{\norm*{b} + \norm*{x_0}}} \nonumber \\
	&+ \norm*{(I - \alpha A)^k} \norm*{x_0 - x^*},\label{E:total_distance}
\end{align}
where \eqref{E:triangle} results from the triangle inequality, and \eqref{E:total_distance} is derived from \eqref{E:distributed_richardson_distance} and \eqref{E:richardson_distance}. In addition to the error by centralized Richardson iteration, the distributed version incurs an error from \eqref{E:distributed_richardson_distance}.
\subsection{Proof of Lemma~\ref{Lem:2}}
\label{Proof:Lem:2}
Using the triangle inequality, we  have
\begin{align*}
	\norm{w_{t} - w^*}
	&= \norm{w_{t-1} - w^* + \eta_{t-1} d_{t-1}^R} \nonumber \\
	&\leq \underbrace{\norm{w_{t-1} - w^* + \eta_{t-1}\hat{d}_{t-1}}}_{T_1} + \eta_{t-1} \underbrace{\norm{ d_{t-1}^R - \hat{d}_{t-1} }}_{T_2}. 
\end{align*}
According to Theorem 4.1 in \cite{polyakNewVersionsNewton2020a}, let: \\
\begin{align}
	t_0 = \max \biggC{0, \bigg \lceil \frac{2L}{\lambda^2\norm{\nabla {f(w_0)}}} \bigg \rceil - 2}, \gamma =  \frac{L}{2\lambda^2}\norm{\nabla {f(w_0)}} - \frac{t_0}{4} \in \big[ 0, \frac{1}{2}\big);  \nonumber
\end{align}
then we have:
\begin{align}
	T_1 \leq \begin{cases}
		\frac{\lambda}{L} (t_0 - t + \frac{2\gamma}{1-\gamma}) , t \leq t_0 \\ \label{E:T1}
		\frac{2\lambda\gamma^{2^{t-t_0}}}{L(1 - \gamma^{2^{t-t_0}})},  t > t_0
		\end{cases}
\end{align}
We next bound $T_2$, which is due to the $\delta$-approximation error. 
	\begin{align*}
		T_2 &= \norm{ d_{t-1}^R - \hat{d}_{t-1} }  = \norm{ \hat{d}_{t-1}  - d_{{t-1},R} + d_{{t-1},R} - d_{t-1}^R }  \\
		& \leq \norm{ \hat{d}_{t-1}  - d_{{t-1},R}} + \norm{d_{{t-1},R} - d_{t-1}^R },
	\end{align*}
where $d_{t-1,R}$ is true Newton direction $d_{t-1}$ obtained by using Richardson iteration after $R$ iteration. From the Theorem \ref{Th:1}, by considering $\hat{d}_{t-1} = x^*$, $d_{{t-1},R} = x_k$, $d_{t-1}^R = \frac{1}{n}\SumLim{i=1}{n}x_{i,k}$, and choosing $\alpha \leq \min \bigC{\frac{1}{R}, \frac{1}{\hat{\lambda}_{max}}}$, we have: 
	\begin{align*}
 \norm{d_{{t-1},R} - d_{t-1}^R }    \leq E_2 = O\biggP{ \frac{\nu}{R^2} \bigP{\norm*{g_{t-1}} + \norm*{d_0}}}
\end{align*}
and
	\begin{align*}
  \norm{ \hat{d}_{t-1}  - d_{{t-1},R}}  \leq \norm*{(I - \alpha A)^R} \norm{ \hat{d}_{t-1} + d_0}, \nonumber \\
\end{align*}
By choosing $ \norm*{d_0} = 0$, we have: 
\begin{align}
		T_2 &\leq \norm*{(I - \alpha A)^R}   \norm{\hat{d}_{t-1} }+ O\biggP{ \frac{\nu}{R^2} \bigP{\norm*{g_{t-1}} }}\nonumber \\
		 &\leq\norm*{(I - \alpha A)^R}   \norm{\hat{d}_{t-1} }  + O\biggP{ \frac{\nu}{R^2} \norm*{H_{t-1}H_{t-1}^{-1}g_{t-1} }}\nonumber \\
		&\leq \norm*{(I - \alpha A)^R}  \norm{\hat{d}_{t-1} } + O\biggP{ \frac{\nu}{R^2}\norm*{H_{{t-1}}}\norm{H_{t}^{-1}g_{t-1}} }\nonumber \\
		&\leq \norm*{(I - \alpha A)^R}    \norm{H_{t-1}^{-1}g_{t-1}}  + O\biggP{ \frac{\nu L}{R^2} \norm{H_{{t-1}}^{-1}g_{t-1}} }\nonumber \\
		&\leq \biggP{\norm*{(I - \alpha A)^R}  + O\BigP{\frac{\nu L}{R^2} }}\norm{H_{t-1}^{-1}g_{t-1}} \nonumber \\
		&\leq  \delta \norm{H_{t-1}^{-1}} \norm{g_{t-1}}  \leq  \frac{\delta}{\lambda} \norm{\nabla f(w_{t-1})},\nonumber
	\end{align}
so
\begin{align*}
\eta_{t-1}T_2	\leq \eta_{t-1} \frac{\delta}{\lambda} \norm{\nabla f(w_{t-1})}, \nonumber
\end{align*}
where 
\begin{align*}
		\delta = \norm*{(I - \alpha A)^R}   + O\BigP{\frac{\nu L}{R^2} },
\end{align*}
In damped phase, when $t \leq t_0$, $\eta_{t-1} = \frac{ \lambda ^2}{L \norm{\nabla {f(w_{t-1})}}}$. In pure Newton phase,  $t >  t_0$ and $\eta_{t-1} = 1$. So 
\begin{align}
	\eta_{t-1}T_2	&\leq
	\begin{cases}
		\frac{\delta\lambda}{L} ,  t \leq t_0 \nonumber\\
		\frac{\delta}{\lambda} \norm{\nabla f(w_{t-1}) },t > t_0  \nonumber\
	\end{cases}\\
	&\leq
	\begin{cases}
		\frac{\delta\lambda}{L} ,  t \leq t_0 \\ \label{E:T2}
		 \delta \kappa \norm{w_{t-1} -w^* },t > t_0 
	\end{cases}
\end{align}
Combine \eqref{E:T1} and \eqref{E:T2} we have: 
\begin{align}
	\norm{w_{t} - w^*} \leq \begin{cases}
		\frac{\lambda}{L} (t_0 - t + \frac{2\gamma}{1-\gamma}) +  \frac{\delta}{\kappa} , t \leq t_0  \label{E:T1T22}\\
		\frac{2\lambda\gamma^{2^{t-t_0}}}{L(1 - \gamma^{2^{t-t_0}})} +   \delta \kappa \norm{w_{t-1} -w^* },  t > t_0 
	\end{cases}
\end{align}
Appling  \eqref{E:T1T22} recursively we have:
\begin{align*}
	\norm{w_{t} - w^*} & \leq \begin{cases}
		\frac{1}{\kappa} (t_0 - t + \frac{2\gamma}{1-\gamma}) +  \frac{\delta}{\kappa} , t \leq t_0 \\ 
		\frac{2t\gamma^{2^{t-t_0}}}{\kappa(1 - \gamma^{2^{t-t_0}})} +  {(\delta\kappa)}^t \norm{w_{0} -w^* },  t > t_0 \nonumber
	\end{cases}
\end{align*}

\subsection{Proof of Theorem \ref{Th:2}}
\label{Proof:Th:2}
From Lemma~\ref{Lem:2}, if $\norm{w_{0} -w^* } \geq \frac{2t\gamma^{2^{t-t_0}}}{\kappa  {(\delta\kappa)}^t(1 - \gamma^{2^{t-t_0}})}$, we have:
\begin{align*}
		\norm{w_{t} - w^*}  \leq  2{(\delta\kappa)}^t \norm{w_{0} -w^* } \leq \epsilon,
\end{align*}
then we can obtain 
\begin{align*}
	T \geq \delta\kappa\frac{\log 2{\norm*{w_{0} - w^*}}/{\epsilon}}{\log {1}/{\gamma}}. 
\end{align*}

\subsection{Proof of Theorem \ref{Th:3}}
\label{Proof:Th:3}
 From Lemma~\ref{Lem:2}, if $\norm{w_{0} -w^* } < \frac{2t\gamma^{2^{t-t_0}}}{\kappa  {(\delta\kappa)}^t(1 - \gamma^{2^{t-t_0}})}$, we have:
\begin{align*}
			\norm{w_{t} - w^*}  \leq  \frac{4t\gamma^{2^{t-t_0}}}{\kappa (1 - \gamma^{2^{t-t_0}})}  \leq \epsilon,
\end{align*}
then we have:
\begin{align*}
 \frac{\kappa (1 - \gamma^{2^{t-t_0}})}{4t\gamma^{2^{t-t_0}}}  & \geq \frac{1}{\epsilon}\\
 \log \frac{\kappa}{4t} + \log(1 - \gamma^{2^{t-t_0}}) + 2^{t-t_0}\log(\gamma)  & \geq \log \frac{1}{\epsilon}
\end{align*}
As $\gamma \in \big[ 0, \frac{1}{2}\big]$, when $t$ increases, $ \log(1 - \gamma^{2^{t-t_0}}) \rightarrow 0$. Besides, $\log \frac{\kappa}{4t}  << 2^{t-t_0}\log(\gamma) $. So, we have:
\begin{align*}
	T = O\BigP{\log \log \frac{1}{\epsilon}}.  
\end{align*}
%
\bibliographystyle{IEEEtran}
\bibliography{DONE}
	\begin{IEEEbiography}[{\includegraphics[width=1in,height=1.25in,clip,keepaspectratio]{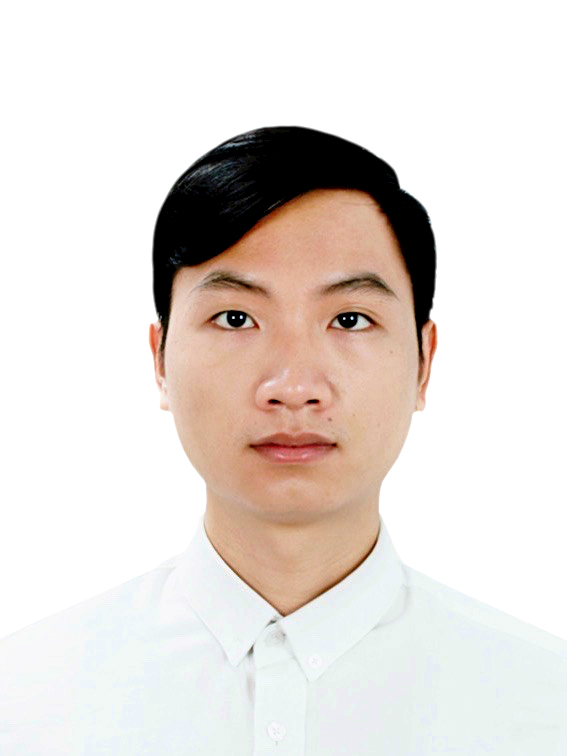}}]
	{\bf Canh T. Dinh} received the BE degree in Electronics and Telecommunication from Ha Noi University of Science and Technology, Ha Noi City, Vietnam, in 2015 and Master of Data Science degree from Université Grenoble Alpes, Grenoble, France, in 2019. He is currently pursuing a Ph.D. degree in Computer Science at The University of Sydney, Sydney, Australia. His supervisor is  Dr. Nguyen H. Tran. His research interests include  Federated Learning and privacy machine learning.
\end{IEEEbiography}
\vskip -2\baselineskip plus -1fil
\begin{IEEEbiography}[{\includegraphics[width=1in,height=1.25in,clip,keepaspectratio]{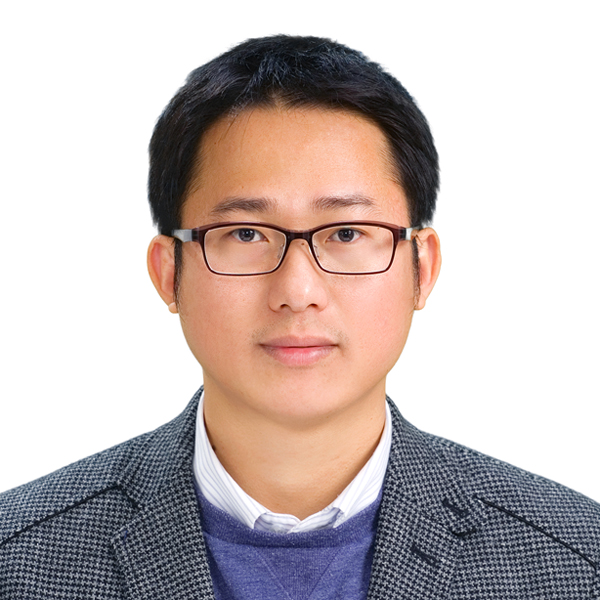}}]
	{\bf Nguyen H. Tran}
	(S\textquoteright10-M\textquoteright11-SM\textquoteright18) received BS and Ph.D degrees, from HCMC University of Technology and Kyung Hee University, in electrical and computer engineering, in 2005 and 2011, respectively. He was an Assistant Professor with Department of Computer Science and Engineering, Kyung Hee University, from 2012 to 2017. Since 2018, he has been with the School of Computer Science, The University of Sydney, where he is currently a Senior Lecturer. His research interests include distributed computing, machine learning, and networking. He received the best KHU thesis award in engineering in 2011 and several best paper awards, including IEEE ICC 2016 and ACM MSWiM 2019. He receives the Korea NRF Funding for Basic Science and Research 2016-2023 and ARC Discovery Project 2020-2023. He was the Editor of IEEE Transactions on Green Communications and Networking from 2016 to 2020, and the Associate Editor of IEEE Journal of Selected Areas in Communications 2020 in the area of distributed machine learning/Federated Learning. 
\end{IEEEbiography}
\vskip -2\baselineskip plus -1fil
\begin{IEEEbiography}[{\includegraphics[width=1in,height=1.25in,clip,keepaspectratio]{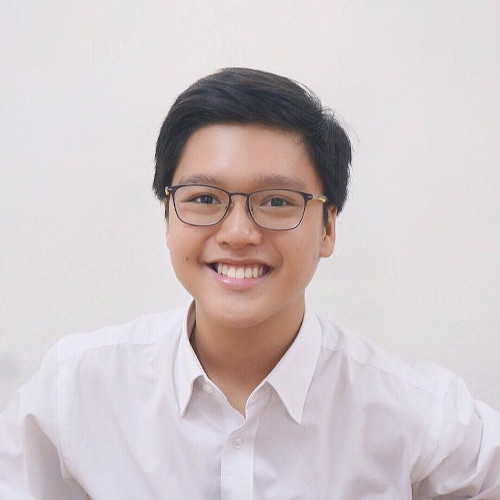}}]
	{\bf Tuan Dung Nguyen}
	received a B.S. in computer science from the University of Melbourne, Australia. He is currently an M.Phil. candidate at the Computational Media Lab, the Australian National University. His research interests include distributed optimization, machine learning and computational social science.
\end{IEEEbiography}
\vskip -2\baselineskip plus -1fil
\begin{IEEEbiography}[{\includegraphics[width=1in,height=1.25in,clip,keepaspectratio]{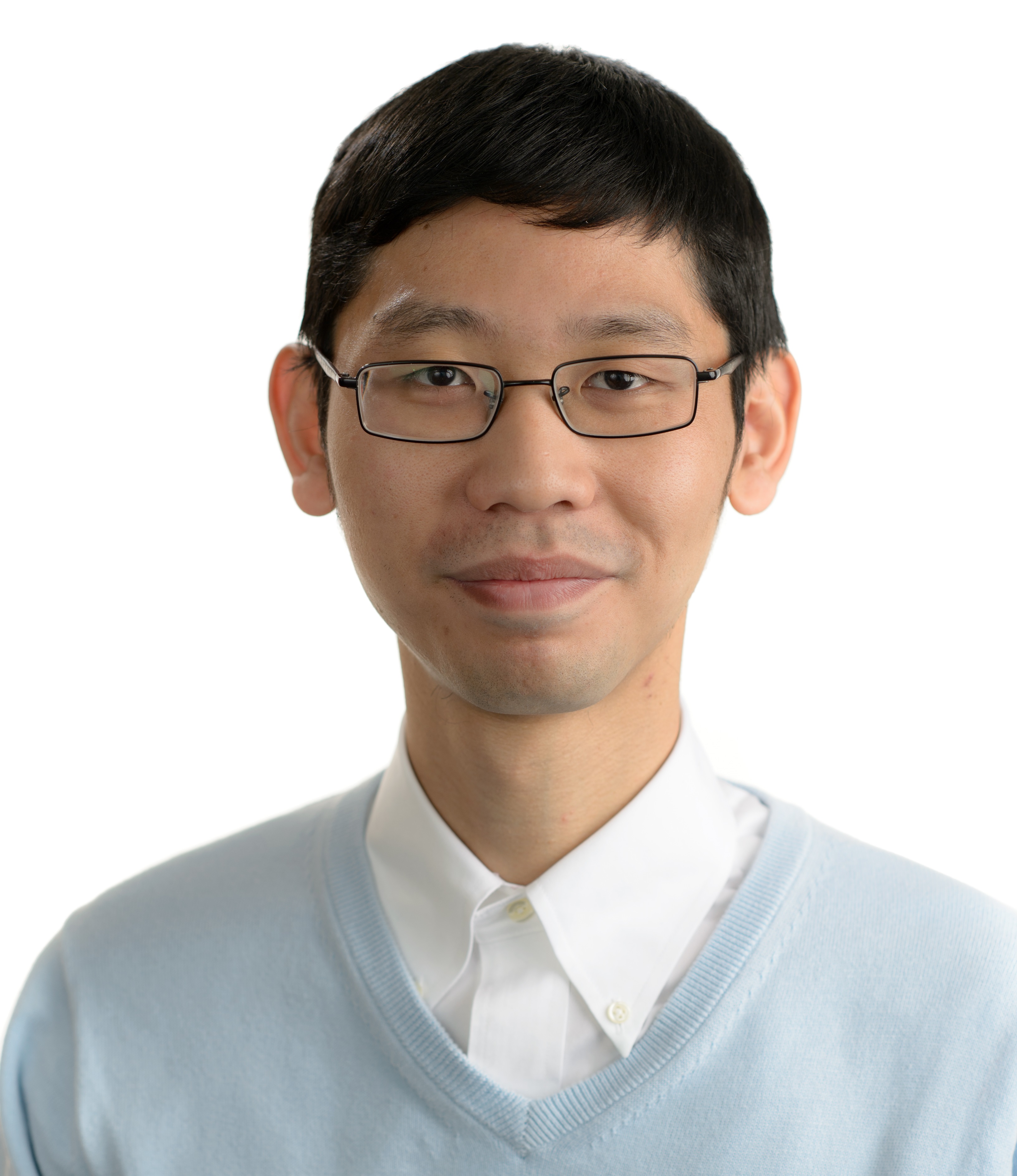}}]
	{\bf Wei Bao}
	(S\textquoteright10-M\textquoteright16) received the B.E. degree in Communications Engineering from the Beijing University of Posts and Telecommunications, Beijing, China, in 2009; the M.A.Sc. degree in Electrical and Computer Engineering from the University of British Columbia, Vancouver, Canada, in 2011; and the PhD degree in Electrical and Computer Engineering from the University of Toronto, Toronto, Canada, in 2016. He is currently a senior lecturer at the School of Computer Science, the University of Sydney, Sydney, Australia. His research covers the area of network science, with particular emphasis on Internet of things, mobile computing, and edge computing. He received the Best Paper Awards in ACM International Conference on Modeling, Analysis and Simulation of Wireless and Mobile Systems (MSWiM) in 2013 and 2019 and IEEE International Symposium on Network Computing and Applications (NCA) in 2016.	\end{IEEEbiography}

\vskip -2\baselineskip plus -1fil
\begin{IEEEbiography}[{\includegraphics[width=1in,height=1.25in,clip,keepaspectratio]{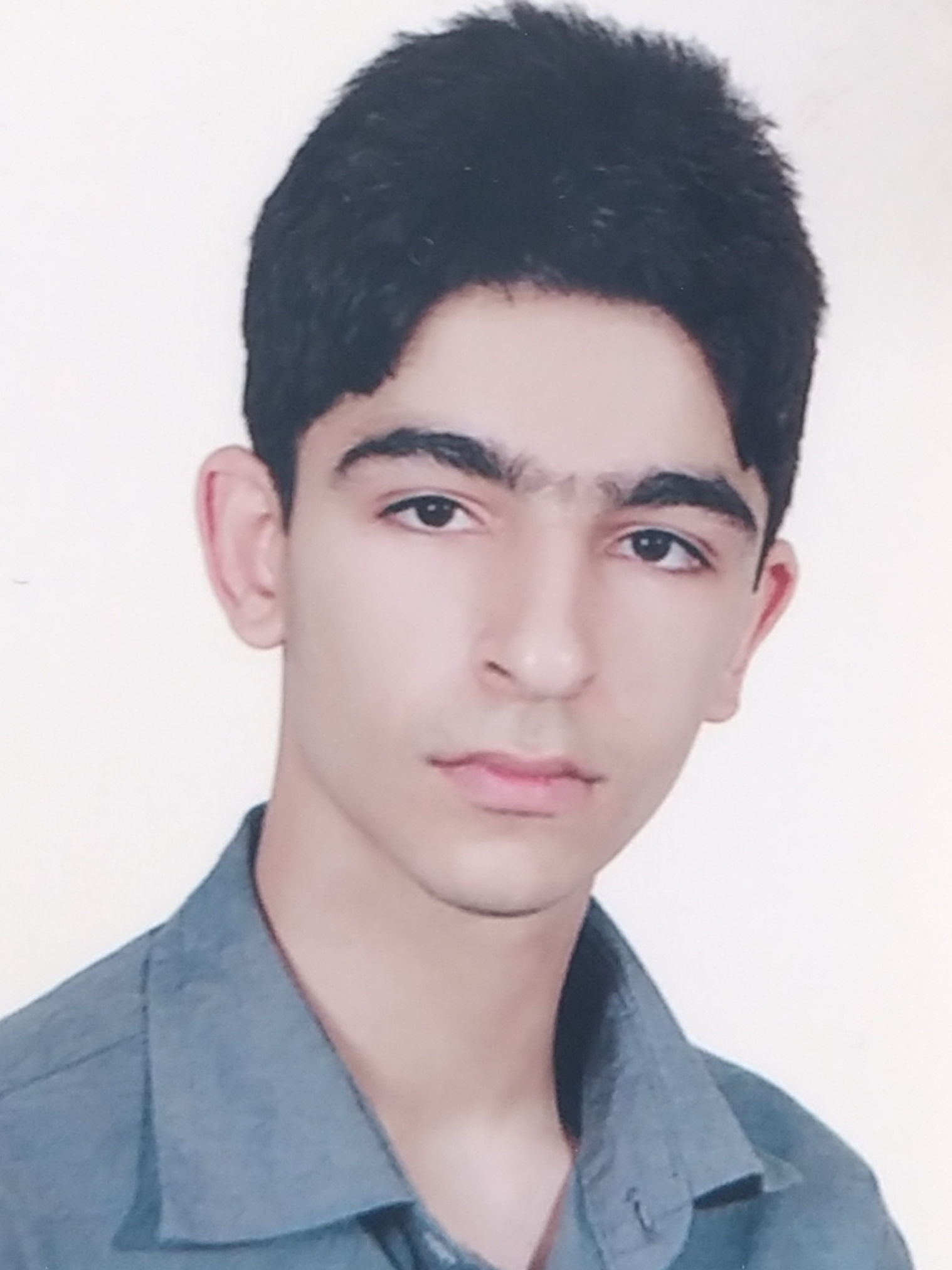}}]
	{\bf Amir R. Balef } received the BE degree in Electrical Engineering from Amirkabir University of Technology, Tehran, Iran, in 2017 and  M.Sc. degree  ,in field of  Digital systems,  from Sharif University of Technology, Tehran, Iran in 2019. His  research  interests  include  edge computing, Internet of Things, privacy-preserving machine learning, federated learning. In recent years, he has focused on optimization algorithms.
\end{IEEEbiography}
\vskip -2\baselineskip plus -1fil
\begin{IEEEbiography}[{\includegraphics[width=1in,height=1.25in,clip,keepaspectratio]{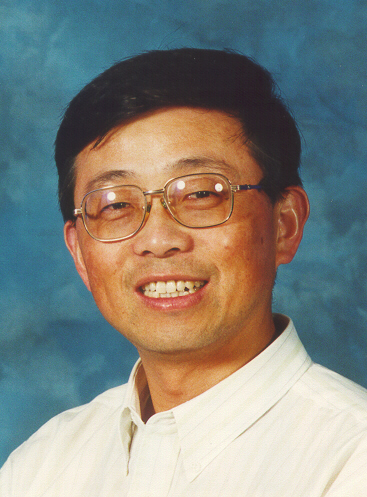}}]
	{\bf Bing Bing Zhou } received the graduate degree in electronic engineering, in 1982 from the Nanjing Institute of Technology in China, and 
the PhD degree in computer science, in 1989 from Australian National University, Australia. 
	He is an associate professor in the School of Computer Science, the University of Sydney, Australia (2003-present). 
	Currently, he is the theme leader for distributed computing applications in the Centre for Distributed and High Performance Computing at the University of Sydney.
\end{IEEEbiography}
\vskip -2\baselineskip plus -1fil
\begin{IEEEbiography}[{\includegraphics[width=1in,height=1.25in,clip,keepaspectratio]{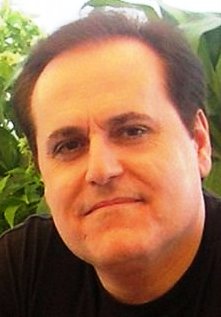}}]
	{\bf Albert Y. ZOMAYA } is Chair Professor of High-Performance Computing and Networking in the School of Computer Science and Director of the Centre for Distributed and High-Performance Computing at the University of Sydney. To date, he has published more than 600 scientific papers and articles and is (co-)author/editor of more than 30 books. A sought-after speaker, he has delivered more than 190 keynote addresses, invited seminars, and media briefings. His research interests span several areas in parallel and distributed computing and complex systems. He is currently the Editor in Chief of the ACM Computing Surveys and served in the past as Editor in Chief of the IEEE Transactions on Computers (2010-2014) and the IEEE Transactions on Sustainable Computing (2016-2020).
	
	Professor Zomaya is a decorated scholar with numerous accolades including Fellowship of the IEEE, the American Association for the Advancement of Science, and the Institution of Engineering and Technology (UK). Also, he is an Elected Fellow of the Royal Society of New South Wales and an Elected Foreign Member of Academia Europaea. He is the recipient of the 1997 Edgeworth David Medal from the Royal Society of New South Wales for outstanding contributions to Australian Science, the IEEE Technical Committee on Parallel Processing Outstanding Service Award (2011), IEEE Technical Committee on Scalable Computing Medal for Excellence in Scalable Computing (2011), IEEE Computer Society Technical Achievement Award (2014), ACM MSWIM Reginald A. Fessenden Award (2017), and the New South Wales Premier’s Prize of Excellence in Engineering and Information and Communications Technology (2019).
\end{IEEEbiography}
\end{document}